\documentclass[pdflatex,sn-mathphys-num]{sn-jnl}

\usepackage{graphicx}%
\usepackage{multirow}%
\usepackage{amsmath,amssymb,amsfonts}%
\usepackage{amsthm}%
\usepackage{mathrsfs}%
\usepackage[title]{appendix}%
\usepackage{xcolor}%
\usepackage{textcomp}%
\usepackage{manyfoot}%
\usepackage{booktabs}%
\usepackage{algpseudocode}%
\usepackage{listings}%

\usepackage{cite}
\usepackage{epstopdf}
\usepackage[ruled]{algorithm2e}
\usepackage{diagbox}
\usepackage{makecell}
\usepackage{dsfont}
\usepackage{subcaption}

\theoremstyle{thmstyleone}%
\newtheorem{theorem}{Theorem}
\newtheorem{lemma}{Lemma}
\newtheorem{corollary}{Corollary}

\theoremstyle{thmstyletwo}%

\theoremstyle{thmstylethree}%
\newtheorem{definition}{Definition}%

\begin{document}

\title[Article Title]{Extended UCB Policies for Multi-Armed Bandit Problems}

\author*[1]{\fnm{Keqin} \sur{Liu}}\email{keqin.liu@xjtlu.edu.cn}


\author[2]{\fnm{Tianshuo} \sur{Zheng}}\email{221501001@smail.nju.edu.cn}

\author[3,4]{\fnm{Zhi-Hua} \sur{Zhou}}\email{zhouzh@nju.edu.cn}

\affil*[1]{\orgdiv{School of Mathematics and Physics}, \orgname{Xi'an Jiaotong-Liverpool University}, \orgaddress{\street{111 Ren'ai Road}, \city{Suzhou}, \postcode{215123}, \state{Jiangsu}, \country{China}}}

\affil[2]{\orgdiv{School of Mathematics}, \orgname{Nanjing University}, \orgaddress{\street{22 Hankou Road}, \city{Nanjing}, \postcode{210093}, \state{Jiangsu}, \country{China}}}

\affil[3]{
\orgname{National Key Laboratory for Novel Software Technology}, \orgaddress{\street{163 Xianlin Road Road}, \city{Nanjing}, \postcode{210023}, \state{Jiangsu}, \country{China}}}

\affil[4]{\orgdiv{School of Artificial Intelligence}, \orgname{Nanjing University}, \orgaddress{\street{163 Xianlin Road}, \city{Nanjing}, \postcode{210023}, \state{Jiangsu}, \country{China}}}


\abstract{The multi-armed bandit (MAB) problems are widely studied in fields of operations research, stochastic optimization, and reinforcement learning. In this paper, we consider the classical MAB model with heavy-tailed reward distributions and introduce the extended robust UCB policy, which is an extension of the results of \citet{5} and \citet{21} that are further based on the pioneering idea of UCB policies [e.g. \citealt{3}]. The previous UCB policies require some strict conditions on reward distributions, which can be difficult to guarantee in practical scenarios. Our extended robust UCB generalizes Lattimore's seminary work (for moments of orders $p=4$ and $q=2$) to arbitrarily chosen $p>q>1$ as long as the two moments have a known controlled relationship, while still achieving the optimal regret growth order $O(\log T)$, thus providing a broadened application area of UCB policies for heavy-tailed reward distributions. Furthermore, we achieve a near-optimal regret order without any knowledge of the reward distributions as long as their $p$-th moments exist for some~$p>1$. Finally, we briefly present our earlier work on light-tailed reward distributions for a complete illustration of the amazing simplicity and power of UCB policies.}

\maketitle

\section{Introduction}
As a prototype of reinforcement learning, a classical multi-armed bandit (MAB) model consists of a player and a bandit machine with~$K$ arms. Under resource constraints, the player can only choose one arm to pull at each time step. The objective of the player is to identify and select the best arm so that the long-term expected reward is maximized. Clearly, the player's action involves balancing the trade-off between exploration and exploitation, i.e., choosing an arm to learn its statistics vs. choosing an arm appearing to be best based on past observations. Specifically, each arm~$i$ offers a random reward according to an unknown distribution $X_i$ when pulled by the player (or agent).  Each reward distribution has a mean, say~$\theta_i$. The player who has limited knowledge about the underlying reward distributions (no knowledge about the means $\{\theta_i\}$), has to choose only one of these arms to play and accrue a reward at each time step, given the reward history so far observed. The player aims to maximize the reward rate that accrues as the time horizon~$T$ grows. In many applications of online decision-making problems, such as Internet advertising and recommendation systems, the MAB model is widely assumed, with numerous algorithms proposed for efficient reinforcement learning, e.g. \citet{37}, \citet{22}. We point out that the way knowledge is acquired becomes crucial to maximizing the long-term reward rate in modern AI systems \citep{30}. Since knowledge can be treated as data in general, the struggle between exploration (sampling) and exploitation (consuming) as a fundamental dilemma is expressly reflected in the MAB model \citep{27, 20, 3, 23, 5}.

To address the key issue in the conflict between exploration and exploitation, we emphasize on the fact that it is impossible for the player to know exactly the specific values of reward means within any finite horizon of time. Instead, the only information the player possesses is the (local) reward history, and the player has to learn from this to make a decision at each time step. Based on the limited historical information, the player has to choose between two specific options. The first option is to explore an apparently suboptimal arm with a lower average reward, as there is always a possibility that the truly best arm may not give the best average reward in the past sample path. The second option is to exploit an arm with the highest average reward offered so far. \citet{27} considered a simple two-armed bandit model and showed that there is a policy to achieve the best average reward asymptotically as time goes to infinity.
Many years later,\citet{20} proposed a much stronger performance measure referred to as regret, which represents the expected total loss compared to the ideal scenario in which the player knows which arm has the highest reward mean from the very beginning.
Given a set of $K$ arms with reward distributions 
$\mathcal F = \{F_1, F_2, \dots, F_K\}$ 
with means 
$\{\theta_i\}_{i=1}^K$ 
and a policy~$\pi$, the regret 
$\mathcal{R}_\pi^\mathcal{F}(T)$ 
is defined as (without loss of generality, we can assume that $\theta_1\ge \theta_i$ for $1\le i\le K$)
\begin{align}\label{eqn:psdoReg}
  \mathcal{R}_\pi^\mathcal{F}(T) = \sum_{\theta_i < \theta_1} \mathbb{E}[s_i(T)] \Delta_i,
\end{align}
where 
\begin{align}\label{def_deltai}
  \Delta_i = \theta_1 - \theta_i,
\end{align} 
and 
$s_i(T)$ 
denotes the number of times that arm $i$ was chosen during time steps 1 to $T$ under the policy $\pi$. Certainly, a lower regret growth rate indicates better performance of a policy and also a higher efficiency of learning. Furthermore, any sublinear regret order with~$T$ implies the asymptotic achievement of the maximum time-average reward as $T\rightarrow\infty$.

Strikingly, \citet{20} proved an asymptotic lower bound $O(\log T)$ on the regret growth rate, 
under the assumption that the probability density/mass function of each reward distribution has the form $f(\cdot; \theta)$, where~$f$ is known but parameter~$\theta$ is unknown.
They also proposed an asymptotically optimal policy for a family of distributions, under which the optimal regret can be achieved asymptotically, both the logarithmic order and the leading coefficient.
\citet{3} proposed a class of policies featured by UCB (upper confidence bound), achieving logarithmic regret if the reward distributions have finite support on a closed interval
$[a, b]\subset\mathbb{R}$ with $a,b\in\mathbb{R}$ known to the player.
Their UCB1 and UCB1-Tuned polices perform very well for the classical MAB problem with a theoretically proven regret bound for any {\em finite} time horizon.
UCB policies are widely applied in machine learning developments (e.g. recommendation systems \citep{31, 16}) and in various reinforcement learning algorithms such as Monte Carlo Tree Search (MCTS) \citep{19}, which is frequently used in gaming AIs \citep{26} and forms a key step in the design of AlphaGo Zero \citep{15}.

\subsection{Light-Tailed and Heavy-Tailed MAB Models}\label{subsec:intro_HT}
In the literature, MAB was first addressed under the assumption that all reward distributions are light-tailed, i.e., the sample mean converges to the true mean faster so learning becomes easier compared to the heavy-tailed case. Specifically, light-tailed (or sub-Gaussian) distributions usually have only a relatively small probability of producing a significantly shifted random sample from its expectation; while heavy-tailed distributions have a relatively high probability of producing a large deviation. There are several popular light-tailed probability distributions, such as Bernoulli, Gaussian, Laplacian, and Exponential. Specifically, the class of light-tailed distributions requires the (local) existence of the moment-generating function of the associated random variable and is therefore referred to as the locally sub-Gaussian distributions \citep{10}.
Formally, a random variable~$X$ is called light-tailed if there exists some $u_0 > 0$ such that its moment-generating function is well defined on $[-u_0,u_0]$ \citep{10}:
\begin{align}\label{eqn:momentGF}
  M(u) := \mathbb{E}[\exp(uX)] < \infty,\ \ \forall |u|\le u_0.
\end{align}
The above condition is equivalent to \citep{10}:
\begin{align}\label{eqn:momentAssume}
  \mathbb{E}[\exp(u(X-\mathbb{E} X))] \le \exp( \zeta u^2 /2),\ \  \forall |u| \le u_0,\ \forall \zeta \ge \sup_{|u|\le u_0} {M^{(2)}(u)},
\end{align}
where $M^{(2)}(\cdot)$ denotes the second derivative of $M(\cdot)$. Note that the above upper bound on $M(u)$ has a form of the moment-generating function of the Gaussian distribution, thus the name ``sub-Gaussian'' has been adopted.

The terminology ``heavy-tailed'', opposite to ``light-tailed'', implies that the reward distribution can produce large values with high probabilities, leading to nonexistence of its moment-generating function (even locally). Compared to the light-tailed class, the heavy-tailed one is harder to learn in terms of the rank of their means. \citet{5} proposed the robust UCB algorithm, under the assumption that there exists a known $p>1$ such that $\mathbb{E}[|X_i|^p]$ exists and is upper bounded by a known parameter, for all $1\le i\le K$. Robust UCB shows significant progress for the heavy-tailed MAB model by achieving the optimal logarithmic order of regret growth. Subsequently, \citet{21} developed a new UCB policy referred to as the \textit{scale free algorithm}. This algorithm removes the general assumption of a known upper bound on the $p$-th moment. Instead, it allows the variance and the squared fourth moment to scale freely as long as their ratio (kurtosis) is bounded by a known parameter.  But this assumption restricts~$p$ to be no less than $4$ (not too heavy-tailed). In this paper, we extend Lattimore's work to any $p>1$ with a tighter regret upper bound in the case of $p=4$ with low discrimination where arm means are close and learning the arm rank becomes challenging.

First, we assume that the player knows a constant $C_{p,q}$ for
some~$p$ and~$q$ ($p>q>1$) such that\footnote{By Jensen's inequality, the existence of the $p$-th moment implies the existence of the $q$-th moment for any $1<q<p$.} 
\begin{align}\label{eqn:momConC}
  \mathbb{E}[|X-\mathbb{E} X|^p] \le C_{p,q} \left[\mathbb{E}[|X-\mathbb{E} X|^q]\right]^{p/q}.
\end{align}
We point out that the knowledge of specific values of~$p$ and~$q$ is not required. Furthermore, the requirement of knowledge of $C_{p,q}$ can also be removed with an arbitrarily small sacrifice of the regret order as seen in Sec.~\ref{sec:extUCB}. The above inequality is a direct generalization of the assumption of kurtosis in \citet{21} where $p=4$ and $q=2$. For a specific example, suppose that~$X$ has a Pareto distribution of type~III with cumulative distribution function \citep{2}
\begin{align*}
  F(x) = \begin{cases}
    1 - \left(1+\left( \frac{x-a}{\sigma} \right)^{1/\gamma}\right)^{-1}, & x>a\\
    0, & x\le a
  \end{cases},
\end{align*}
where $0 < \gamma < 1$ and $\sigma>0$. Simple computations show that
\begin{align*}
  \mathbb{E}[|X-\mathbb{E} X|^p] = C(p, q, \gamma) \left[\mathbb{E}[|X-\mathbb{E} X|^q]\right]^{p/q}
\end{align*}
for $1<q<p<1/\gamma$, where 
\begin{align*}
  I(p, \gamma) &:= \int_{0}^{+\infty} \left(x-\frac{\gamma \pi}{\sin \gamma \pi}\right)^p \frac{x^{1/\gamma}}{\gamma x (1+x^{1/\gamma})^2}\,{\rm d}x,\\
  C(p, q, \gamma) &= I(p, \gamma)/ I(q, \gamma)^{p/q}.
\end{align*}
Note that the moment of this distribution is a function of~$\sigma$. When~$\sigma$ increases, the moment of any fixed order also increases if it exists. The robust UCB policy requires the knowledge of~$p$ and an upper bound of the $p$-th moment \citep{5}. 
Therefore, for an arm with an unknown~$\sigma$, we cannot apply the robust UCB policy. 
Furthermore, learning efficiency can be improved if a larger~$p$ can be used (tighter bound on the tail, but the corresponding moment bound becomes harder to know or compute {\it a priori}). It is thus better to eliminate the need for any prior knowledge regarding the specific values or bounds on the moments. This motivates \citet{21} to consider the scale free algorithm. However, Lattimore's algorithm requires specific values of~$p$ and~$q$ ($p=2q=4$) and cannot deal with the very heavy-tailed case (for $p$ close to~$1$). Such requirements significantly limited the application of the scale free algorithm. In contrast, our algorithm only requires $p>1$ while~$p$ can be unknown or unspecified. Furthermore, our algorithm achieves near-optimal regret even without any specific knowledge on the reward distributions including $C_{p,q}$. These results significantly broaden the applicability of the scale free algorithm. After we generalize Lattimore's work, we will give a correction to a minor flaw of a lemma in \citet{5} that subsequently caused a small error in the computation of the regret upper bound in \citet{21}.

\subsection{Related Work}\label{sec:relatedWork}
The theory of frequentist MAB has been developed for a long period of time, and various milestones have been made since \citet{20}. In contrast to the Bayesian model, the frequentist MAB does not assume a priori knowledge of the initial probability or state of the system but learns the core parameters solely through the past sampling history, e.g., computing an upper confidence bound as a function of the frequency that an arm is selected and the values of the observed samples. This way of decision-making in the balance of exploration with exploitation contrasts with the Bayesian model in which the tradeoff is usually addressed by dynamic programming for the transitional characteristics of the system or the observation model \citep{12}. 
After the UCB1 policy by \citet{3}, many other policies of the UCB class have been proposed in the literature. \citet{25} directed their attention to the prospect of KL divergence between probability distributions. They introduced the KL-UCB algorithm, which notably tightened the regret bound. This improvement was achieved under the assumption of bounded rewards with known bounds. Furthermore, KL-UCB policy achieves the asymptotic lower bound of regret growth rate for this class of rewards given in \citet{6}. \citet{18} combined the Bayesian method with the UCB class and proposed Bayes-UCB for bounded rewards with known bounds, achieving a theoretical regret bound similar to that of KL-UCB. Numerical experiments demonstrated the high efficiency of learning by Bayes-UCB for Gaussian reward distributions. A survey of such UCB policies can be found in \citet{7}. These policies were neither implemented nor proved to achieve the logarithmic regret growth in the case of unbounded reward distributions.
In 2011, the first author of this paper proposed UCB1-LT with a proven logarithmic upper bound on the regret growth rate for the class of all light-tailed reward distributions \citep{23}, filling the gap for the case of unbounded light-tailed reward distributions. Together with the extended robust UCB for the heavy-tailed class focused in this paper, we provide a relatively complete picture of UCB policies for general reward distributions. We will first present our UCB policy for the heavy-tailed class as the main content of this paper, followed by a brief presentation of UCB1-LT. These results demonstrate the simplicity and strong performance of UCB for all reward distributions. The UCB for the heavy-tailed case is of course more complicated because the player needs to extract more information from past observation in contrast to the simple form (e.g. sample mean) of UCB as in the light-tailed case. After 2011, \citet{4} extended UCB1-LT to 
$(\alpha, \psi)$-UCB under a more general assumption that the (light-tailed) reward moment-generating function is bounded by a convex function~$\psi$ for all $u\in \mathbb R$.

For the case of heavy-tailed reward distributions, we have mentioned that \citet{5} proposed the robust UCB policy under the assumption that an upper bound of the moments is known. The robust UCB policy uses different mean estimators rather than the empirical one and remarkably achieves the logarithmic regret growth rate. Under certain regularity conditions, the robust UCB policy may adopt three different mean estimators: the truncated mean estimator, the median-of-means estimator, and the Catoni mean estimator \citep{8}.
The truncated mean estimator requires the knowledge of an upper bound on the origin moments, which is not very effective in the case that the reward mean is far away from zero. The median-of-means estimator considers the (central) moment and thus patches up the deficiency of the truncated mean estimator: the mean estimation will not be affected when the reward distributions are transformed by translations. The Catoni mean estimator is more complicated and can only be used when the reward distributions have variances ($p\ge2$). But when applied in the robust UCB policy, it gives a much smaller regret coefficient than the other two estimators. Recently, \citet{11} showed the result of applying the Catoni mean estimator to the case of $p<2$, but the performance was not proven to be better than the truncated mean and the median-of-means estimators. Finer estimations when extending the Catoni mean estimator to the case of $p<2$ are interesting for future investigations. Several recent papers also considered the heavy-tailed MAB model and made significant progress. \citet{35} proposed the robust MOSS policy with logarithmic regret growth. \citet{1} established a policy called $\text{KL}_{\inf}$-UCB, which achieves a logarithmic upper bound on regret very close to the asymptotic lower bound proved by \citet{6}. For other non-classical MAB models, the heavy-tailed model has also been considered \citep{36,29}. However, these policies do not eliminate the knowledge requirement of an upper bound of moments, which has been removed from the extended robust UCB policy proposed in this paper. Note that \citet{38} considered the case where no knowledge of the moments is available except for the existence of moments of order greater than~$1$, but a relatively strong assumption of the distribution (a larger probability mass on the negative semi-axis) was needed to achieve logarithmic regret.

There are also policies without following the idea of UCB. The $\epsilon$-greedy policy was well studied, and many policies were proposed using its idea, e.g., the constant $\epsilon$-decreasing policy~\citep{33}, GreedyMix~\citep{9}, and $\epsilon_n$-greedy~\citep{3}. The Deterministic Sequencing of Exploration and Exploitation (DSEE) policy~\citep{24,32} also achieves logarithmic regret for both cases of single player and multiple players (with knowledge of upper bound on moments). Other recent progresses on variants of MAB, such as adversarial bandits, contextual bandits, and linear bandits, can be found in \citep{22,13,17}. In the following table, we list the main results found in the literature for classical MAB with heavy-tailed reward distributions for comparisons with ours.

\resizebox{\textwidth}{!}
{
\begin{tabular}{|l|c|c|}
\hline
\diagbox[width=4.5cm, height=1.2cm]{Papers}{Results} & Logarithmic Regret Bound & Assumption\\
\hline 
\citet{5} & $\sum_{i:\Delta_i > 0} \left( \frac{v}{\Delta_i} \right)^{1/\epsilon}\log T$ & $\mathbb{E}[|X_i-\theta_i|^{1+\epsilon}] \leq v~\text{(known)}, ~\forall i$\\
\hline
\citet{21} & $\sum_{i:\Delta_i > 0} \left( (\kappa-1)\Delta_i+\frac{\sigma_i^2}{\Delta_i} \right)\log T $& $\mathrm{Kurt}[X_i] \leq \kappa~\text{(known)}, ~\forall i,~\sigma_i:=\sqrt{\mathrm{Var}(X_i)}$\\
\hline
\citet{35} & $\sum_{i:\Delta_i > 0} \log\left( \frac{T\Delta_i^{\frac{1+\epsilon}{\epsilon}}}{K}\right)\frac{1}{\Delta_i^{1/\epsilon}}$ & $\mathbb{E}[|X_i-\theta_i|^{1+\epsilon}] \leq v~\text{(known)}, ~\forall i$\\
\hline
\citet{1}$^{\dagger}$ & $\sum_{i:\Delta_i > 0} \frac{\log T}{KL_{\inf} (\mathcal{F}_i,\theta_1)}$ & $\mathbb{E}[|X_i-\theta_i|^{1+\epsilon}] \leq v~\text{(known)}, ~\forall i$\\
\hline
\citet{38} & $ \sum_{i:\Delta_i > 0} \left(\left(\frac{v}{\Delta_i}\right)^{1/\epsilon}+\frac{\Delta_i}{\mathbb{P}(X_i \neq 0)}\right)\log T $& $\mathbb{E} [X_1 \mathds{1}_{\{|X_1| > M \}}] \leq 0, ~\forall M \geq 0$\\
\hline
Ours & \makecell{$\sum_{i:\Delta_i > 0}  \Delta_i\Big(144(1+C_{p,q})(6v_q^{\frac{1}{q}}+2^q)(\Delta_i+\frac{1}{\Delta_i})\Big)^{M(p,q)}\log T$, \\ $where \quad M(p,q)= 1/[(1-\frac{1}{q})(1-\frac{q}{p})]>1 $} &$v_p/v_q^{p/q} \leq C_{p,q}~\text{(known)}, ~1<q<p, ~\forall i,~v_l:=\mathbb{E}[|X_i-\theta_i|^l] $\\ 
\hline  
& & \makecell{
$^{\dagger} KL_{\inf}(\eta,x):=\inf\{KL(\eta,\mathcal{F}):\mathbb{E}[|X|^{1+\epsilon}] \le v\quad and \quad \mathbb{E}_{X\sim\mathcal{F}}[X] \ge x\}$}\\
\hline
\end{tabular}
}

\section{Heavy-Tailed Reward Distributions}
\subsection{The Extended Robust UCB Policy}\label{sec:extUCB}
Consider~$K$ arms offering random rewards $\{X_1, X_2, \dots, X_K\}$ with heavy-tailed distributions $\mathcal F = \{F_1, F_2, \dots, F_K\}$. In contrast with the light-tailed class, a heavy-tailed distribution may yield a very high (or low) reward realization with a high probability. Consequently, the empirical mean estimator $\overline X_{i,s}=\frac 1s \sum_{j=1}^s X_{i,j}$ for a sequence of i.i.d. random variables
$\{X_{i,j}\}_{j=1}^s$
may not work properly to estimate the expectation of $X_i$ for arm~$i$. The appendix of \citet{5} has shown that under the assumption of heavy-tailed distributions, with probability at least $1-\delta$,
$$
  \frac 1s \sum_{j=1}^s X_{i,j} \le 
  \theta_i + \left(
    \frac{3 \mathbb{E}|X_i-\theta_i|^r}{\delta s^{r-1}}
  \right)^{\frac{1}{r}},
$$
where $1<r\le 2$. However, this bound is too loose to obtain a UCB policy for achieving logarithmic regret growth. To remedy this issue, we adopt the median-of-means estimator proposed in \citet{5}.



\begin{definition}\label{medianofmeanDef}
  For a finite i.i.d. sequence of random variables
  $\{X_{i,j}\}_{j=1}^s$ 
  drawn from a distribution 
  $F_i$, 
  define the median-of-means estimator 
  $\hat \mu(\{X_{i,j}\}_{j=1}^s, k)$ with 
  $k$ $(k\le s)$
  bins as {\bf the median of 
  $k$ 
  empirical means}
  \begin{align*}
    \left\{ \frac{1}{N} \sum_{j=lN+1}^{(l+1)N} X_{i,j} \right\}_{l=0}^{k-1},
  \end{align*}
  where $N = \lfloor s/k \rfloor$. 
\end{definition}


The following lemma offers an approximation of the proximity between the median-of-means estimator, utilizing a specific number of bins, and the actual mean.
It should be noted that this lemma is similar to Lemma 2 in \citet{5}. 
However, there is a minor flaw in the proof of the latter, which will be explained later. Moreover, there exist differences between the two lemmas, prompting us to restate and establish their distinctions in the following exposition.
\begin{lemma}\label{lemma_medianofmean_probability}
  Given any $\epsilon>0$. Suppose that 
  \begin{align*}
    \log \delta^{-1}>1/\epsilon,\ s\ge (8+\epsilon) \log \delta^{-1} \frac{8\log ( \delta ^{-1})+1}{\epsilon\log ( \delta ^{-1})-1}
  \end{align*}
  and 
  $v = \mathbb{E}[|X_i - \theta_i|^p] < \infty$ for some $p$ such that $p>1$.
  Then with probability at least $1-\delta$, the median-of-means estimator
  \begin{align*}
    \hat \mu(\{X_{i,j}\}_{j=1}^s, \lceil 8 \log \delta^{-1} \rceil) \ge \theta_i - (12 v)^{1/p} \left(
      \frac{(8+\epsilon) \log(\delta ^{-1})}{s}
    \right)^{(p-1)/p}.
  \end{align*}
  Also, with probability at least $1-\delta$,
  \begin{align*}
    \hat \mu(\{X_{i,j}\}_{j=1}^s, \lceil 8 \log \delta^{-1} \rceil) \le \theta_i + (12 v)^{1/p} \left(
      \frac{(8+\epsilon) \log(\delta ^{-1})}{s}
    \right)^{(p-1)/p}.
  \end{align*}
  \end{lemma}

With the help of the median-of-means estimator, we can now define the upper confidence bound for the extended robust UCB policy.
As mentioned in~\eqref{eqn:momConC}, we first assume that for any arm distribution 
$X_i \in \mathcal F$ with mean $\theta_i$, 
there is a known moment control coefficient
$C_{p,q}$ ($1<q<p$)
such that
\begin{align}\label{moment_control_coef_def}
  \mathbb{E}[|X_i-\theta_i|^p] \le C_{p,q} [\mathbb{E}[|X_i-\theta_i|^q]]^{r},
\end{align}
where $r = p/q$.

\begin{definition}\label{def_ucb_of_extendedRobustUCB}
 Let $k = \lceil 8\log \delta^{-1} \rceil,\chi =  \frac{(8+\epsilon) \log \delta^{-1}}{s}$. Given an arbitrary $\epsilon>0$, a moment order $p>1$, another moment order $1<q<p$, and the moment control coefficient $C_{p,q}$, define the upper confidence bound of the extended robust UCB as (value of the fraction is set to $+\infty$ if divided by $0$)
\begin{align*}
  &\tilde \mu(\{X_{i,j}\}_{j=1}^s, \delta):= \\
  &\sup
  \left\{\rule{0cm}{1cm}
    \theta \in \mathbb R:\ \theta \le \hat\mu(\{X_{i,j}\}_{j=1}^s, k) + \left(
    \frac{12 \hat\mu(\{|X_{i,j}-\theta|^q\}_{j=1}^s, k)}{
      \max\left\{ 0, 1-  C'\chi^{(p-q)/p} \right\}
    }
  \right)^{1/q} \chi^{(q-1)/q} 
  \rule{0cm}{1cm}\right\},
\end{align*}
where 
\begin{align}\label{short_notes_C}
  C' = \left(12(C_{p,q} + 1)\right)^{q/p}.
\end{align}
\end{definition}

Note that this upper confidence bound always exists, since the left-hand side of the above inequality tends to~$-\infty$ as $\theta \to -\infty$, while the right-hand side tends to $+\infty$.


\begin{lemma}\label{lemma_mutilde_optimal}
  Suppose that the conditions in Lemma~\ref{lemma_medianofmean_probability} are satisfied, then with probability at least 
  $1-2\delta$, 
  we have
  \begin{align*}
    \tilde \mu(\{X_{i,j}\}_{j=1}^s, \delta) \ge \theta_i.
  \end{align*}
\end{lemma}

\begin{corollary}\label{corollary_medianofmean}
Under the conditions of Lemma~\ref{lemma_medianofmean_probability}, with probability at least 
$1-\delta$, 
we have
\begin{align*}
   \hat\mu(\{Y_{i,j}\}_{j=1}^s, k) - v_q \le C' v_q\chi^{(p-q)/p}.
\end{align*}
\end{corollary}
Please see~\eqref{yij_def} for the definitions of $Y_{i,j},v_q,Y_i,v_p$. Finally, with the upper confidence bound $\tilde \mu(\{X_{i,j}\}_{j=1}^s) $, we complete the extended robust UCB policy as follows.

\begin{algorithm}[H]
  \caption{The Extended Robust UCB Policy~$\pi^*$}
  \SetKwInOut{KIN}{$a_1 \ge 8\zeta$, $a_2 \ge a_1/(\zeta u_0)$, $K$}
  \text{Input: $\epsilon>0$, $K$: the number of arms}\\
  \text{Initialize: $s_i\leftarrow 0$ and $\{X_{i,j}\}_{j=1}^{s_i}$ an empty sequence for each arm $i$, $1\le i \le K$}\\
  \For{$t \leftarrow 1$ to $T$}{
    \For{$i \leftarrow 1$ to $K$}{
      \If{$s_i=0$}{Set arm~$i$'s upper confidence bound to $+\infty$}
      \Else{Compute arm~$i$'s upper confidence bound as $\tilde \mu(\{X_{i,j}\}_{j=1}^{s_i}, t^{-2})$}
    }
    Choose an arm $j$ that maximizes the upper confidence bound and obtain reward $x$\\
    Update: $s_j\leftarrow s_j+1$ and append $x$ to sequence $\left\{X_{j,k}\right\}_{k=1}^{s_j}$
  }
\end{algorithm}


\vspace{0.5cm}
Here we present the main theorem that the extended robust UCB policy achieves the logarithmic order of regret growth. 

\begin{theorem}\label{theorem_extended_robust_UCB}
  Suppose $p >q> 1$ such that the $p$-th order moments exist for all $F_i \in \mathcal F$ and \eqref{moment_control_coef_def} is satisfied with known $C_{p,q}$.
  Then for any $\epsilon>0$, the regret $\mathcal{R}_{\pi^*}^\mathcal{F}(T)$ of the extended robust UCB policy~$\pi^*$ has the logarithmic order with respect to the time horizon~$T$. Specifically, we have $\mathcal{R}_{\pi^*}^\mathcal{F}(T)\le \sum_{i:\Delta_i > 0} C_i(T)\Delta_i$ with
  \begin{align*}
  C_i(T) := \Big[36(1+\frac{8}{\epsilon})+(16+2\epsilon)\Big(144(1+C_{p,q})(6v_q^{\frac{1}{q}}+2^q)(\Delta_i+\frac{1}{\Delta_i})\Big)^{M(p,q)}\Big]\log T  + 11,
\end{align*}
where $M(p,q) := 1/((1-\frac{1}{q})(1-\frac{q}{p}))$ and $T \geq \frac{9}{2\epsilon}+\frac{3}{2}.$
\end{theorem}


From the above, we observe that the upper bound on regret increases with $C_{p,q}$ but decreases with~$p$. This matches the intuition that a tighter known bound on the moment ratio or a higher moment order adopted makes the learning process easier due to less variation assumed in the samples. Furthermore, for any fixed $q$, the upper bound increases with $v_q$. This is also easy to explain since a larger moment indicates a larger deviation from a sample to the true mean. Meanwhile, if~$q$ increases and becomes close to $p$, then the denominator of $M(p,q)$ will approach~$0$ and the upper bound becomes worse. This is because a larger value of~$p/q$ offers more global information on moments with different orders to facilitate the learning of the unknown distributions. Also, as $\Delta_i\rightarrow0$ for some~$i$, the upper bound goes to infinity (since $M(p,q)>1$) because it will be harder to distinguish the best arm from arm~$i$. Finally, if the arm number $K$ is increased with other parameters including~$\Delta_i$ relatively fixed within the original orders, the upper bound is clearly increased linearly with~$K$ due to the definition of regret. To prove Theorem~\ref{theorem_extended_robust_UCB}, we just need to show the following core lemma to bound the time~$s_i$ spent on each suboptimal arm $i$ by the extended robust UCB policy.

\begin{lemma} \label{lemma_chosen_time_estimate_extended_robust}
  The expected number of times $\mathbb{E}[s_i(T)]$ that a suboptimal arm is chosen by the extended robust UCB policy~$\pi^*$ has an upper bound of the logarithmic order with~$T$. Specifically, we have $\mathbb{E}[s_i(T)]\le C_i(T)$ where $C_i(T)$ is defined above.
\end{lemma}

\begin{figure}[htbp]
    \centering
    \begin{subfigure}[t]{0.45\textwidth}
        \centering
        \includegraphics[width=\textwidth]{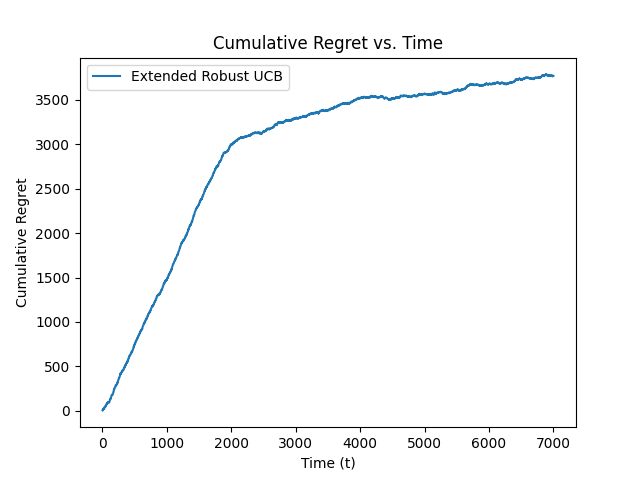}
        \caption{Laplace with $C_{p,q}=1.68\times 10^3$}
        \label{fig:LA_q3p9}
    \end{subfigure}
    \hfill
    \begin{subfigure}[t]{0.45\textwidth}
        \centering
        \includegraphics[width=\textwidth]{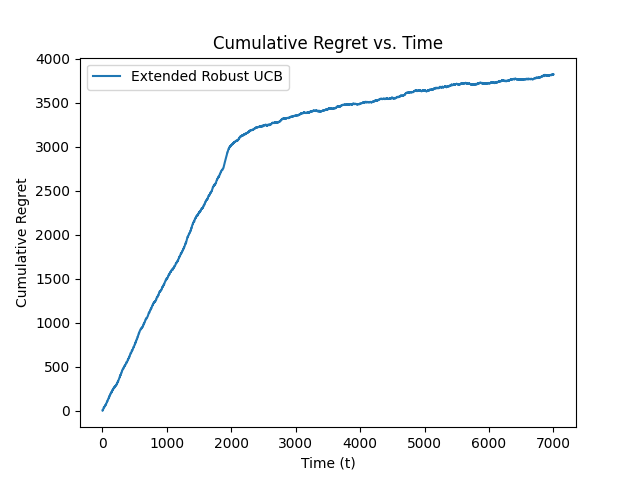}
        \caption{Student-t with $C_{p,q}=2.09\times 10^3$}
        \label{fig:SA_q3p9_K2}
    \end{subfigure}
    \caption{Comparison for different~$C_{p,q}$}
    \label{fig:Cpq comparison}
\end{figure}

\begin{figure}[htbp]
    \centering
    \begin{subfigure}[b]{0.45\textwidth}
        \centering
        \includegraphics[width=\textwidth]{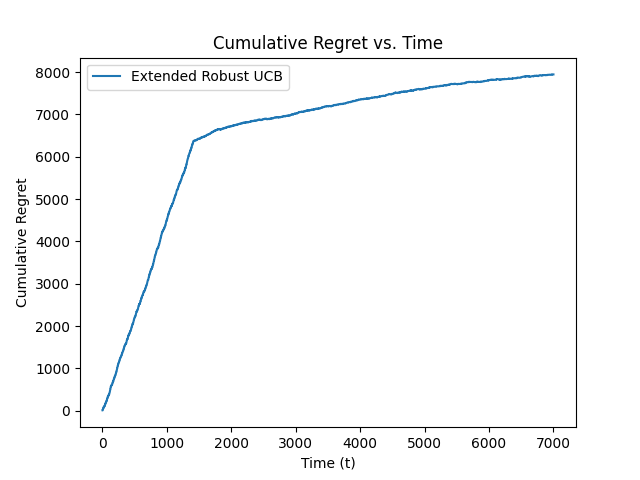}
        \caption{Pareto III with $p=12$}
        \label{fig:PA_q4p12}
    \end{subfigure}
    \hfill
    \begin{subfigure}[b]{0.45\textwidth}
        \centering
        \includegraphics[width=\textwidth]{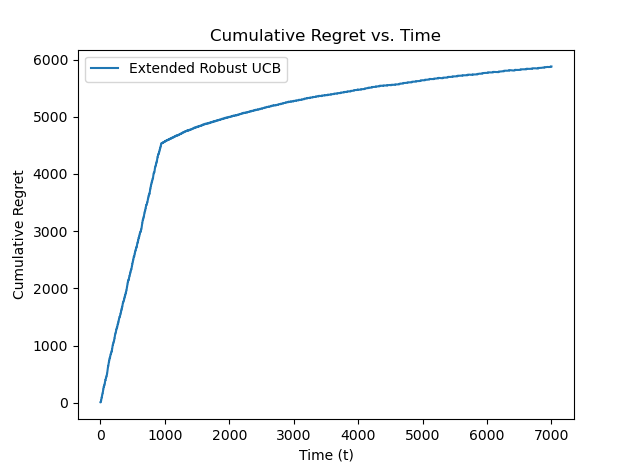}
        \caption{Pareto III with $p=15$}
        \label{fig:PA_q4p15}
    \end{subfigure}
    \caption{Comparison for different~$p$}
    \label{fig:p comparison}
\end{figure}

\begin{figure}[htbp]
    \centering
    \begin{subfigure}[b]{0.45\textwidth}
        \centering
        \includegraphics[width=\textwidth]{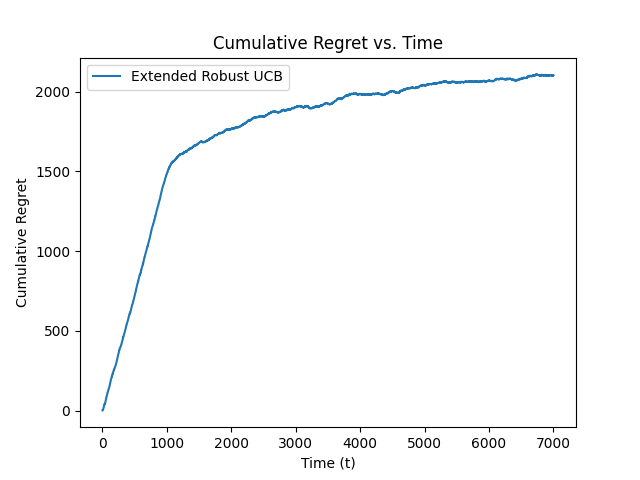}
        \caption{Laplace with $v_q=24$}
        \label{fig:LA_q4p15_lowVq}
    \end{subfigure}
    \hfill
    \begin{subfigure}[b]{0.45\textwidth}
        \centering
        \includegraphics[width=\textwidth]{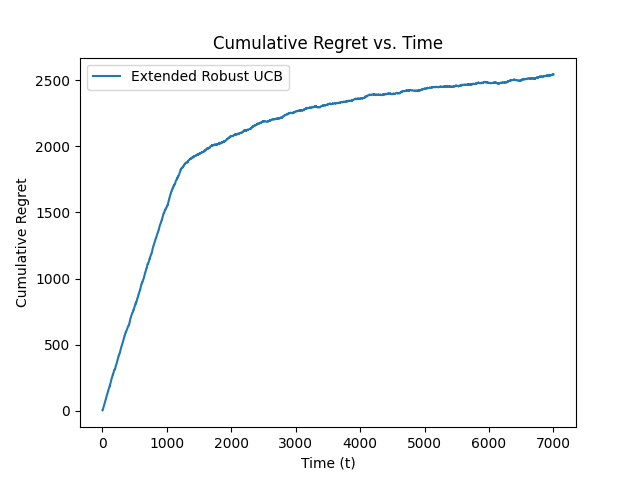}
        \caption{Laplace with $v_q=384$}
        \label{fig:LA_q4p15_highVq}
    \end{subfigure}
    \caption{Comparison for different~$v_{q}$}
    \label{fig:Vq comparison}
\end{figure}

\begin{figure}[htbp]
    \centering
    \begin{subfigure}[b]{0.45\textwidth}
        \centering
        \includegraphics[width=\textwidth]{figs/SA_q3p9_K2.png}
        \caption{Student-t with $K=2$}
        \label{fig:SA_q3p9_K2}
    \end{subfigure}
    \hfill
    \begin{subfigure}[b]{0.45\textwidth}
        \centering
        \includegraphics[width=\textwidth]{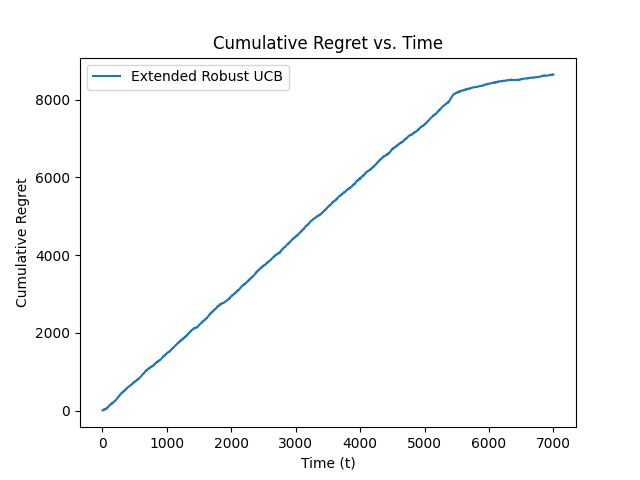}
        \caption{Student-t with $K=4$}
        \label{fig:SA_q3p9_K4}
    \end{subfigure}
    \caption{Comparison for different~$K$}
    \label{fig:K comparison}
\end{figure}

Finally, we conduct Monte Carlo simulation examples based on Laplace distributions, Student-t distributions, and Pareto distributions of type~III (see Sec.~\ref{subsec:intro_HT}). From Fig.~\ref{fig:Cpq comparison}--\ref{fig:K comparison}, we observe that the regret rate grows linearly at the beginning to prepare the ground for ranking until at least one arm has a finite UCB (i.e., $\chi$ defined in~\eqref{short_notes_chi} becomes sufficiently small such that, within the $\sup$ operator of $\tilde \mu(\{X_{i,j}\}_{j=1}^{s_i}, t^{-2})$, not only the denominator is nonzero but also the growth rate of the right-hand side with $\theta$ becomes small). In Fig.~\ref{fig:Cpq comparison}, we compare the actual regret for different values of $C_{p,q}$ and observe that the regret becomes larger for larger $C_{p,q}$. In Fig~\ref{fig:p comparison}, the regret becomes smaller as~$p$ becomes larger. Furthermore, the regret increases with~$v_q$ and~$K$ as shown in Fig.~\ref{fig:Vq comparison} and Fig.~\ref{fig:K comparison}, respectively. All these observations match our theoretical analysis on the upper bound above.

Note that if $C_{p,q}$ is unknown, then the player can simply choose any increasing function~$f$ with $f(t) \rightarrow +\infty$ as $t \rightarrow +\infty$ to replace $C_{p,q}$ and achieve a regret arbitrarily close to logarithmic order without requiring any knowledge of the reward moments (see Appendix~\ref{pf_thm_2} for a proof).


\subsection{Further Analysis and Comparison of The Extended Robust UCB}

Now we consider the specific chosen values of~$p$ and~$q$ to further improve the efficiency of the extended robust UCB policy.

{\it Case~1.} Suppose that the $p'$-th order of moments of reward distributions exists with $p'>2$ for all arms. Then we can choose~$p$ and $q$ such that $p = 2q\le p'$. Inequality~\eqref{C_deflate} can be refined to
\begin{align}
  \begin{split}\label{C_deflate_refine}
    \mathbb{E} [|Y_i - v_q|^{p/q}] = \mathbb{E} Y_i^2 + v_q^2 - 2 v_q\mathbb{E}  Y_i = \mathbb{E} Y_i^2 - v_q^2
  = v_p -v_q^2 \le (C_{p,q}-1) v_q^{p/q}.
  \end{split}
\end{align}
Hence, we can refine $C'$ from \eqref{short_notes_C} to
\begin{align}
  C' = \sqrt{12 (C_{p,q}-1)},
\end{align}
which can be much smaller than the original definition \eqref{short_notes_C}.

Next, we look at the definition of $\chi_M$ in Appendix~\ref{pf_thm_1} that guarantees the three inequalities~\eqref{ineq_1}, \eqref{ineq_2} and~\eqref{ineq_3} to hold. The refinement of $C'$ leads to a smaller right-hand side for all these inequalities, i.e. the constant $\chi_M$ will be larger and the regret upper bound will be smaller, which further implies that the policy's selection will converge to the best arm faster.

{\it Case~2.} Assume that the $4$-th order of moments of reward distributions exists as in \citet{21}. Then we can choose $p=4$ and $q=2$.
In this case, our UCB in Definition~\ref{def_ucb_of_extendedRobustUCB} can be rewritten as
\begin{align*}
  \tilde \mu(\{X_{i,j}\}_{j=1}^s, \delta):=\sup
  \left\{\rule{0cm}{1cm}
    \theta \in \mathbb R:\ \theta \le \hat\mu(\{X_{i,j}\}_{j=1}^s, k) + \sqrt{
    \frac{12 \hat\mu(\{(X_{i,j}-\theta)^2\}_{j=1}^s, k)}{
      \max\left\{ 0^+, 1-  C'\sqrt\chi \right\}
    }
    \chi}
  \rule{0cm}{1cm}\right\}.
\end{align*}
In other words, $C_{p,q}$ degenerates to an upper bound on kurtosis.
The function $B(x)$ defined in~\eqref{def_bchi} becomes
\begin{align*}
  B(x) = \frac{24\tau(x)}{1-48x\tau(x) }\left(
    1 + C' \sqrt{x}
  \right) + \frac{576 \sqrt{x}\tau(x) }{1-48 x\tau(x) },
\end{align*}
and the three inequalities \eqref{ineq_1}, \eqref{ineq_2} and \eqref{ineq_3} are refined to
\begin{align}
  1 & > C' \sqrt{x},\label{modified_cond_1}\\
  1& > 48 x\tau(x) ,\label{modified_cond_2}\\
  \Delta_i & \ge \sqrt{v_2 x}\left(
    \sqrt{B(x)}+2\sqrt 3
  \right).\label{modified_cond_3}
\end{align}

Under these refined estimations for $\chi_M$, the leading coefficient of $\log T$ in \eqref{our_regret_of_HT} is significantly reduced. Now we compare the extended robust UCB and the scale free algorithm proposed in \citet{21}, under the same assumption specified at the beginning of Case~2.

We first address a flaw of Lemma~2 in \citet{5}. This lemma, crucial in \citet{21} for deriving the regret upper bound of the scale free algorithm, is examined more closely below. For clarity and coherence with the notation employed in this paper, we present the following adapted version.

\noindent {\it A Flawed Statement in \citet{5}}:  Let $\delta \in (0,1)$ and $p \in (1, 2]$. Let $\{X_{i,j}\}_{j=1}^s$ be i.i.d. random variables with mean $\mathbb{E} X_i = \theta_i$ and (centered) $p$-th moment~$v_p$. Let $k' = \lfloor \min\{8 \log ({\rm e}^{1/8}\delta^{-1}), s/2\}\rfloor$.
  Then with probability at least $1-\delta$,
  \begin{align*}
    \hat \mu\left(
      \{X_{i,j}\}_{j=1}^s, k'
    \right) \le \theta_i + (12 v_p)^{1/p} \left(
      \frac{16 \log({\rm e}^{1/8}\delta^{-1})}{s}
    \right)^{(p-1)/p}.
  \end{align*}

The last inequality in the proof of this statement in \citet{5} assumes that $\exp(-k'/8) \le \delta$, equivalent to $k' \ge 8 \log \delta^{-1}$. 
However, the definition of $k'$ in this statement is
\begin{align*}
  k' = \lfloor \min\{8 \log({\rm e}^{1/8} \delta^{-1}), s/2\}\rfloor.
\end{align*}
In the following, we give the corrected version of the above statement.
\begin{lemma}\label{correction_lemma_2_ucbht}
  Let $\delta \in (0,1)$ and $p \in (1, 2]$. Let $\{X_{i,j}\}_{j=1}^s$ be i.i.d. random variables with mean $\mathbb{E} X_i = \theta_i$ and (centered) $p$-th moment~$v_p$, where~$s$ is chosen such that $\lfloor s/2\rfloor \ge 8 \log \delta^{-1}$. Let $k' = \lfloor \min\{8 \log ({\rm e}^{1/8}\delta^{-1}), s/2\}\rfloor$.
  Then with probability at least $1-\delta$,
  \begin{align*}
    \hat \mu\left(
      \{X_{i,j}\}_{j=1}^s, k'
    \right) \le \theta_i + (12 v_p)^{1/p} \left(
      \frac{16 \log({\rm e}^{1/8}\delta^{-1})}{s}
    \right)^{(p-1)/p}.
  \end{align*}
\end{lemma}


The next theorem shows that with a properly chosen parameter $\epsilon>0$, the upper bound given in~\eqref{our_regret_of_HT} is also tighter than that in \citet{21} in long run if $\Delta_i$ is small, i.e., arms are hard to be distinguished (the low-discrimination case). In the following theorem, we formally state this result while correcting the upper bound in \citet{21} by Lemma~\ref{correction_lemma_2_ucbht}.
\begin{theorem}\label{theorem_compare_scalefree}
  Choose any 
  $8<\epsilon<\frac{280-16 \sqrt{2}}{2 \sqrt{2}+3}$ 
  ($\approx 44.158$)
  and let $\delta = t^{-2}$. 
  The upper bound given in~\citet{21} of $\mathbb{E}[s_i(T)]$ is greater than that given in \eqref{our_regret_of_HT} for sufficiently large $T$ and sufficiently small $\Delta_i$. That is, after dividing both regret upper bounds by $\log \delta^{-1}$ and letting $T\to +\infty$, we have
  \begin{align}
    (8+\epsilon) \left(\frac8\epsilon + \max\left\{ \frac8\epsilon, \frac 1{\chi_M} \right\}\right)
    <
    3648 \max\left\{\frac{(C')^2}{12}, \frac{v_2}{\Delta_i^2}\right\} + 16
  \end{align}
  for sufficiently small $\Delta_i$.
  The regret upper bound of the scale free algorithm proposed in \citet{21} is thus asymptotically larger than that of the extended robust UCB policy for $\epsilon \in(8, \frac{280-16 \sqrt{2}}{2 \sqrt{2}+3})$.
\end{theorem}

Furthermore, we show that the regret of the extended robust UCB deviates from the theoretical lower bound (in the non-parametric setting) by only a constant factor and a constant term as in the scale free algorithm of \citet{21}.
First, we restate the lower bound derived in \citet{21} as follows.
\begin{theorem}\label{theorem_theoretical_bound}
  Let $H_{\kappa_0}$ be the set of distributions that have kurtosis less than $\kappa_0$.
  Assume $\Delta > 0$ and $\kappa_0 \ge 7/2$. Suppose that $X \in H_{\kappa_0}$ has a mean of $\mu$, a positive variance of $\sigma^2$, and a kurtosis of $k$.
  Then
  \begin{align}
    \begin{split}
      \inf\{{\rm KL}(X,X'):& X' \in H_{\kappa_0}\ \text{and} \ \mathbb{E} [X'] > \mu + \Delta\}\\
      \le
      &\begin{cases}
        \min \{ -\log(1-p), \frac{C_1 \Delta^2}{\sigma^2} \}, & \text{if} \ C_0 \sqrt{k} (k+1) \frac{\Delta}{\sigma} < \kappa_0\\
        - \log (1-p), & \text{otherwise}
      \end{cases},
    \end{split}
  \end{align}
  where $C_0$, $C_1 > 0$ are universal constants and $p = \min\{ \Delta / \sigma,\  1/ \kappa_0 \}$.
\end{theorem}
Based on this, we can draw the following conclusion:
\begin{theorem}\label{theorem_lower_bound}
  Let $\delta = t^{-2}$. Under Case~2, the regret of the extended robust UCB policy applied to $H_{\kappa_0}$ ($\kappa_0 \ge 7/2$) differs from the lower bound of regret by a constant factor and a constant term.
\end{theorem}

Furthermore, we give another upper bound that shows a sublinear regret in terms of arm number~$K$. Recall the upper bound on regret in Theorem~\ref{theorem_extended_robust_UCB}. We have
\begin{align*}
\mathcal{R}_{\pi^*}^\mathcal{F}(T) & = \sum_{i:\Delta_i > 0} \Delta_i \mathbb{E}[s_i(T)] = \sum_{i:\Delta_i > 0} \Delta_i [\mathbb{E}[s_i(T)]]^{1/M(p,q)}[\mathbb{E}[s_i(T)]]^{1-1/M(p,q)} \\
&\le \sum_{i:\Delta_i > 0} \Delta_i  \Big[\Big(M_1+M_2(p,q)(\frac{\Delta_i^2+1}{\Delta_i})^{M(p,q)}\Big)2\log T \Big]^{1/M(p,q)}[\mathbb{E}[s_i(T)]]^{1-1/M(p,q)}\\
&\le \sum_{i:\Delta_i > 0}  M_3(p,q)(\Delta_i^2+1)(\log T)^{1/M(p,q)}[\mathbb{E}[s_i(T)]]^{1-1/M(p,q)}\\
&\le M_3(p,q)(\log T)^{1/M(p,q)}\Big[\sum_{i:1\ge\Delta_i > 0}2[\mathbb{E}[s_i(T)]]^{1-1/M(p,q)}+\\&~~~~~~~~~~~~~~~~~~~~~~~~~~~~~~~~~~~~~~~~~~~~~~~~~~~~~~~~~~~~~\sum_{i:\Delta_i  >1}2\Delta_i^2[\mathbb{E}[s_i(T)]]^{1-1/M(p,q)}\Big]\\
&\le 2M_3(p,q)(\log T)^{1/M(p,q)}\Big[K^{1/M(p,q)}T^{1-1/M(p,q)}+\sum_{i:\Delta_i  >1}\Delta_i^2T^{1-1/M(p,q)}\Big]\\
&\le M_3(p,q)(\log T)^{1/M(p,q)}T^{1-1/M(p,q)}\Big[K^{1/M(p,q)}+\sum_{i:\Delta_i  >1}\Delta_i^2\Big]
\end{align*}
where $M_1=19+144/\epsilon,~M(p,q) = 1/((1-1/q)(1-q/p))>1,~T \geq \frac{9}{2\epsilon}+\frac{3}{2}$, and
\begin{align*}
M_2(p,q)&=(8+\epsilon)\Big(144(1+C_{p,q})(6v_q^{\frac{1}{q}}+2^q)\Big)^{M(p,q)}, \\
M_3(p,q)&= \Big(M_1+2M_2(p,q)\Big)^{1/M(p,q)}.
\end{align*}
It is desirable to remove the dependency of regret bound on~$\Delta_i$ as well and this will be considered in the future work.


\section{Light-Tailed Reward Distributions}
We point out that our policy and results also work for the light-tailed reward distributions since our construction depends only on the existence of moments. However, as mentioned in Sec.~\ref{sec:relatedWork}, the UCB could be significantly simplified for the light-tailed class where learning reward means becomes much easier. Furthermore, the leading constant of the logarithmic regret upper bound can also be greatly reduced as shown in \citep{23}. Therefore, we now formally but briefly discuss the work in \citep{23} that extended the UCB policy for distributions with known finite supports in \citep{3} to all light-tailed distributions. We will also compare our work with some subsequent work following \citep{23}. Together with above results for the heavy-tailed class, we see a complete picture of the beauty and power of UCB as well as what information needs to be extracted in its specific design based on how the samples represent the true means.

\subsection{The UCB1-LT Policy}
The empirical mean is adopted in estimating the upper confidence bound for the class of light-tailed reward distributions. As before, let 
$\overline X_{i,s}$ denote the empirical mean 
$\frac1{s}\sum_{j=1}^{s} X_{i,j}$ 
for arm~$i$, 
where 
$\{X_{i,j}\}_{j=1}^s$ form the i.i.d. random reward sequence drawn from an unknown distribution 
$X_i \in \mathcal F$. 

\begin{lemma}\label{Bernstein_type_bound}
(Bernstein-type bound) For i.i.d. random variables 
$\{X_{i,j}\}_{j=1}^s$ drawn from a light-tailed distribution $X_i$ (with mean $\theta_i$)
with a finite moment-generating function 
$M(u)$ 
over range 
$u\in [-u_0, u_0]$ for some $u_0>0$.
We have, 
$\forall \epsilon > 0$,
\begin{align}\label{eqn:BernsteinBound}
  \mathbb P(\overline X_{i,s} - \theta_i \ge \epsilon) \le 
  \begin{cases}
    \exp(-\frac{s}{2\zeta} \epsilon^2), & \epsilon < \zeta u_0\\
    \exp(-\frac{su_0}{2} \epsilon), & \epsilon \ge \zeta u_0\\
  \end{cases},
\end{align}
where 
$\zeta>0$ satisfies $\zeta\ge\sup_{|u|\le u_0} {M^{(2)}(u)}$.
A similar bound also holds for 
$\mathbb P(\overline X_{i,s} - \theta_i \le -\epsilon)$ 
by symmetry.
\end{lemma}

The proof of the above lemma follows a similar argument as in \citet{34}.
Using the Bernstein-type bound by Lemma~\ref{Bernstein_type_bound}, we propose the UCB1-LT policy as follows.

\begin{algorithm}[H]
  \caption{The UCB1-LT Policy}
  \text{Input: $a_1 \ge 8\zeta$, $a_2 \ge a_1/(\zeta u_0)$, $K$: the number of arms}\\
  \text{Initialize: $s_i \leftarrow 0$ and $\overline{X}_i\leftarrow 0$ for each arm $i$, $1\le i \le K$}\\
  \For{$t \leftarrow 1$ to $T$}{
    \For{$i \leftarrow 1$ to $K$}{
      \If{$s_i=0$}{Assign $+\infty$ to the upper confidence bound}
      \Else{
        \If{$\sqrt{\frac{a_1 \log t}{s_i}}<\zeta u_0$}{
          Compute the upper confidence bound as $\overline{X}_{i} + \sqrt{\frac{a_1 \log t}{s_i}}$
        }
        \Else{
          Compute the upper confidence bound as $\overline{X}_{i} + \frac{a_2 \log t}{s_i}$
        }
      }
    }
    Choose an arm $j$ that maximizes the upper confidence bound and obtain reward $x$\\
    $\overline X_{j}\leftarrow \frac{x + s_j \overline X_j}{s_j+1}$ and $s_j \leftarrow s_j+1$
  }
\end{algorithm}

\vspace{0.5cm}

The UCB1-LT policy considers two upper confidence bounds and alternatively uses one of them according to values of $t$ and $s_i(t-1)$. To minimize the theoretical regret upper bound, we can choose $a_1 = 8\zeta$ and $a_2 = \frac{8}{u_0}$. The following theorem shows that UCB1-LT achieves logarithmic regret growth for the light-tailed reward distributions.

\begin{theorem}\label{theorem_ucb1lt}
For the light-tailed reward distributions, the regret of the UCB1-LT policy satisfies the following inequality:
\begin{align}\label{eqn:regLT}
  \mathcal{R}_\pi^\mathcal{F}(T) \le \sum_{i:\theta_i < \theta_1} \Delta_i \left(\max\left\{\frac{4a_1}{\Delta_i^2}, \frac{2a_2}{\Delta_i}\right\} \log T + 1 + \frac{\pi^2}{3}\right).
\end{align}
\end{theorem}
In Appendix~\ref{sec:numUCBLT}, we compare our UCB1-LT policy with the $(\alpha, \psi)$-UCB policy proposed in \citet{4}.


\section{Conclusion and Acknowledgment}

In this paper, we have proposed two order-optimal UCB policies, namely the extended robust UCB and UCB1-LT, dealing with the heavy-tailed and light-tailed reward distributions in the frequentist multi-armed bandit problems, respectively. We are grateful for the help from Mr. Haoran Chen during the initial stage of this project, and Dr. Yaoqing Yang from Dartmouth College in improving this article.



\newpage
\appendix
\section{From UCB1-LT to $(\alpha, \psi)$-UCB}\label{sec:numUCBLT}

After the establishment of UCB1-LT by~\citet{23}, ~\citet{4} subsequently proposed the $(\alpha, \psi)$-UCB under a more general assumption on the moment generating functions. 
Specifically, the 
$(\alpha, \psi)$-UCB policy
assumes that the distribution of reward 
$X_i \in \mathcal F$ 
satisfies the following condition: there exists a convex function $\psi$ defined on 
$\mathbb R$ 
such that for all 
$\lambda > 0$,
\begin{align}\label{alpha_psi_assumption}
  \log \mathbb{E}\left[{\rm e}^{\lambda (X_i - \theta_i)}\right] \le \psi(\lambda) 
  \text{ and }
  \log \mathbb{E}\left[{\rm e}^{\lambda ( \theta_i-X_i )}\right] \le \psi(\lambda).
\end{align}
For example, if $X_i$ is bounded in $[0, 1]$ as assumed in UCB1, one can choose
$\psi(\lambda) = \frac{\lambda^2}{8}$ 
and \eqref{alpha_psi_assumption} becomes the well-known Hoeffding's lemma \citep{4}. Since $+\infty$ is allowed in the range of $\psi$, this assumption is more general than that in UCB1-LT.
The Legendre-Frenchel transform $\psi^*$ of $\psi$ is defined as
\begin{align*}
  \psi^*(\epsilon) := \sup_{\lambda \in \mathbb R} \lambda \epsilon - \psi(\lambda).
\end{align*}
The upper confidence bound of $(\alpha, \psi)$-UCB at time $t$ is defined as
\begin{align}\label{alpha_psi_ucb_def}
  I_i(t) := \overline X_{i,s_i(t-1)} + (\psi^*)^{-1}\left(
    \frac{\alpha \log t}{s_i(t-1)}
  \right).
\end{align}
\begin{theorem}[\citet{4}]\label{alpha_psi_theorem}
    Assume that the reward distributions satisfy \eqref{alpha_psi_assumption}. Then $(\alpha, \psi)$-UCB with $\alpha>2$ achieves
  \begin{align}\label{eqn:alphaUCBreg}
    \mathcal{R}_\pi^\mathcal{F}(T) \le \sum_{\theta_i<\theta_1} \left(
      \frac{\alpha \Delta_i}{\psi^*(\Delta_i/2)}\log T + \frac{\alpha}{\alpha-2}
    \right).
  \end{align}
\end{theorem}

For the UCB1-LT policy, we can choose 
\begin{align}\label{alpha_psi_psi_assumption}
  \psi(\lambda) = \begin{cases}
    \frac{\zeta \lambda^2}{2}, & \lambda \le u_0\\
    +\infty ,& \lambda > u_0
  \end{cases}.
\end{align}
Then \eqref{alpha_psi_assumption} becomes the same assumption as in UCB1-LT (see~\eqref{eqn:momentAssume}). In this case, we have
\begin{align*}
  \psi^*(\epsilon)  = \begin{cases}
    \frac{\epsilon^2}{2\zeta}, & |\epsilon| < \zeta u_0\\
    u_0(|\epsilon| - \frac{\zeta u_0}2), & |\epsilon|\ge \zeta u_0
  \end{cases}.
\end{align*}
Thus the inverse $(\psi^*)^{-1}(x)$ of $\psi^*$ for $x\ge 0$ will be 
\begin{align*}
  (\psi^*)^{-1}(x) = \begin{cases}
    \sqrt{2\zeta x},& 0\le x<\frac{\zeta u_0^2 }2\\
    \frac{x}{u_0} + \frac{\zeta u_0}{2},&x\ge \frac{\zeta u_0^2 }2
  \end{cases}.
\end{align*}
Furthermore, if we choose $\alpha = 4$, then the upper confidence bound defined in \eqref{alpha_psi_ucb_def} becomes
\begin{align}\label{alpha_psi_ucb_def_in_LT}
  I_i(t) = \overline X_{i,s_i(t-1)} + \begin{cases}
    \sqrt{\frac{8\zeta \log t}{s_i(t-1)}},&\frac{\log t}{s_i(t-1)}< \frac{\zeta u_0^2}{8}\\
    \frac{4\log t}{u_0 s_i(t-1)}+\frac{\zeta u_0}2, & \frac{\log t}{s_i(t-1)}\ge \frac{\zeta u_0^2}{8}
  \end{cases}.
\end{align}
If we choose $a_1 = 8\zeta$ and $a_2 = \frac{8}{u_0}$ in UCB1-LT, we get the same UCB under the condition $\frac{\log t}{s_i(t-1)} < \frac{\zeta u_0^2}{8}$.
For the case $\frac{\log t}{s_i(t-1)} \ge \frac{\zeta u_0^2}{8}$, the upper confidence bound of $(\alpha, \psi)$-UCB is not larger than that of UCB1-LT and thus may achieve a smaller theoretical upper bound of regret as follows:
\begin{align}\label{result_alpha_psi}
  \mathcal{R}_\pi^\mathcal{F}(T) \le
  \sum_{\theta_i<\theta_1} \left(
    r(\Delta_i)\log T + 2
  \right),
\end{align}
where
\begin{align*}
  r(\Delta_i) = \begin{cases}
    \frac{32 \zeta}{\Delta_i}, & \Delta_i < 2\zeta u_0\\
    \frac{8\Delta_i}{u_0 (\Delta_i - \zeta u_0)},&\Delta_i \ge 2\zeta u_0
  \end{cases}.
\end{align*}
Note that this regret bound is better than~\eqref{eqn:regLT} in Theorem~\ref{theorem_ucb1lt} for UCB1-LT as $T$ becomes sufficiently large, as shown in the following lemma. 
\begin{lemma}\label{regret_ub_comp_lt}
  Choose the parameter $\alpha$ in $(\alpha, \psi)$-UCB such that $\alpha \le 4$, and assume that Assumption~\eqref{eqn:momentAssume} is satisfied and thus $\psi(\lambda)$ is chosen as in~\eqref{alpha_psi_psi_assumption}. Then the logarithmic regret upper bound in Theorem~\ref{alpha_psi_theorem} has a leading constant no larger than that in Theorem~\ref{theorem_ucb1lt}, i.e.
  \begin{align*}
    \sum_{\theta_i<\theta_1} \frac{\alpha \Delta_i}{\psi^*(\Delta_i/2)} \le  \sum_{\theta_i<\theta_1} \Delta_i \max\left\{\frac{4a_1}{\Delta_i^2}, \frac{2a_2}{\Delta_i}\right\}.
  \end{align*}
\end{lemma}

\begin{figure}[h]

    \begin{minipage}{0.5\linewidth}
        \centerline{\includegraphics[width=\linewidth]{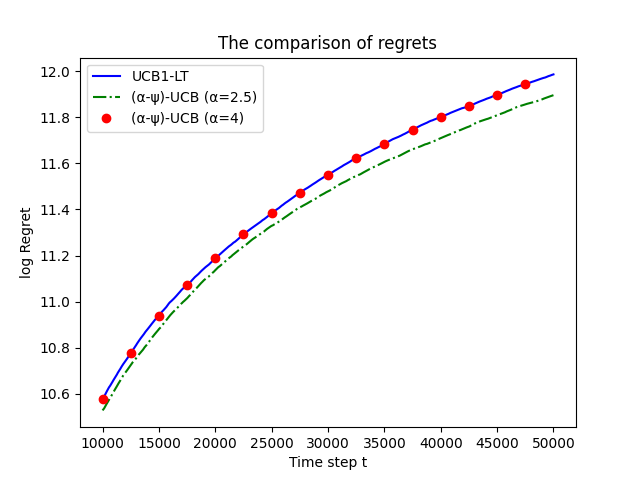}}    
    \end{minipage}
    \begin{minipage}{0.5\linewidth}
        \centerline{\includegraphics[width=\linewidth]{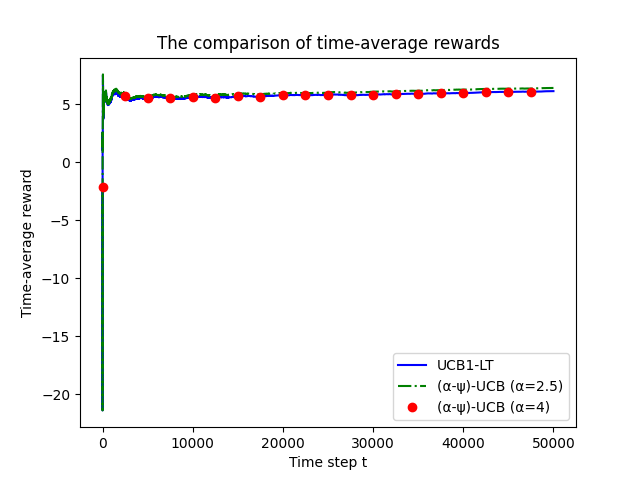}}    
    \end{minipage}

    \caption{Performance comparison between UCB1-LT and $(\alpha, \psi)$-UCB}
    \label{lt_regret_vs_to_alphapsi}
\end{figure}

Now we compare the actual performance between $(\alpha, \psi)$-UCB and UCB1-LT through numerical experiments shown in Fig.~\ref{lt_regret_vs_to_alphapsi}.
Assume that the arms are offering rewards according to~$19$ normal distributions.
The means of reward distributions are $0.5, 1.0, 1.5, \dots, 9.5$ with standard deviations all set to~$60$.
A Monte Carlo (MC) simulation with 100 runs and time period $T=50000$ is shown below.
The horizontal axis in both figures is the time step~$t$.
The first figure only shows~$T$ from $10000$ to $50000$ for better display, while the second one ranges from~$1$ up to $50000$.
The vertical axis of the left figure is the logarithmic regret averaged from the 100 MC runs (to approximate the expectation), while of the right one is the approximately expected time-average reward.
The parameters for UCB1-LT and $(\alpha, \psi)$-UCB are respectively $u_0=1$ and $\alpha = 2.5$ or $4$ with $\zeta=3600$ as their common parameter.

These results reveal an interesting phenomenon: the actual performance of UCB-LT is similar to that of $(\alpha, \psi)$-UCB which has a better theoretical regret bound over long-run for $\alpha=4$.
However, a smaller choice of $\alpha$ slightly improves the performance of $(\alpha, \psi)$-UCB over long-run as consistent with the theoretical bound in \eqref{eqn:alphaUCBreg} whose leading constant of the logarithmic order decreases as $\alpha$ decreases.

\section{Proof of Lemma~\ref{lemma_medianofmean_probability}}\label{sec:proof_lemma_1}
   Observe that $s\ge \lceil 8 \log \delta^{-1} \rceil$ holds under the given condition in the lemma.
    For simplicity of presentation, we introduce some symbols as below.
    Let
  \begin{align}
    k &:= \lceil 8\log \delta^{-1} \rceil,\label{short_notes_k}\\
    \chi &:=  \frac{(8+\epsilon) \log \delta^{-1}}{s},\label{short_notes_chi}\\
    \eta &:= (12v)^{1/p} \chi^{(p-1)/p},\\
    \hat\mu_l &:= \frac{1}{N} \sum_{j=lN+1}^{(l+1)N} X_{i,j}, \ \ l = 0,1, \dots, k-1.
  \end{align}
  According to the appendix of \citet{5},
  \begin{align*}
    \xi := \mathbb P\left( \hat\mu_l > \theta_i + \eta \right) \le \frac{3v}{N^{p-1}\eta^p} = \frac1{4 N ^{p-1}\chi^{p-1}}.
  \end{align*}
  Note that 
  $N = \lfloor s/ \lceil 8\log \delta^{-1} \rceil \rfloor$.
  Since 
  $s\ge (8+\epsilon) \log \delta^{-1} \frac{8\log ( \delta ^{-1})+1}{\epsilon\log ( \delta ^{-1})-1}$, 
  a direct computation yields 
  $\xi \le 1/4$.
  Then using Hoeffding's inequality for the tail of a binomial distribution \citep{14},
  we have
  \begin{align*}
    \mathbb P\left( \hat \mu(\{X_{i,j}\}_{j=1}^s, k) > \theta_i + \eta \right) 
    \le \mathbb P\left(
      \sum_{l=0}^{k-1} \mathbb{I} (\hat \mu_l \ge \theta_i + \eta)\geq \frac k2
    \right)
    \le \exp(-2k (1/2-\xi)^2) \le \delta.
  \end{align*}
  Similarly, the other inequality also holds by symmetry.
  $\hfill\qedsymbol$

\section{Proof of Lemma~\ref{lemma_mutilde_optimal}}\label{sec:proof_of_lemma_2}
Recall the notations introduced by~\eqref{short_notes_k}, \eqref{short_notes_chi} and \eqref{short_notes_C}.
If 
$\tilde \mu(\{X_{i,j}\}_{j=1}^s, \delta) = +\infty$, 
the inequality holds with probability $1$. 
Assume that 
$\tilde \mu(\{X_{i,j}\}_{j=1}^s, \delta)<+\infty$, 
i.e.
\[
  1- C' \chi^{(p-q)/p}>0.
\] 
Define
\begin{align}
  Y_i &:= |X_i - \theta_i|^q, \notag \\
  Y_{i,j} &:= |X_{i,j} - \theta_i|^q,\label{yij_def}\\
  v_q &:= \mathbb{E} Y_i, \notag \\ 
  v_p &:= \mathbb{E} [|X_i - \theta_i|^p]. \notag
\end{align}
From Lemma~\ref{lemma_medianofmean_probability}, we have
\begin{align}
  \begin{split}\label{probability_two_variances_order}
  \mathbb P  & \left(\rule{0cm}{0.8cm}
    \frac{\theta_i - \hat\mu(\{X_{i,j}\}_{j=1}^s, k)}{(12v_q)^{1/q}}
    \le \chi^{(q-1)/q}\right.\\
    &\ \text{and} \  \frac{v_q - \hat\mu(\{Y_{i,j}\}_{j=1}^s, k)}{\left(12 \mathbb{E} [\left|
    Y_i - v_q
    \right|^{p/q}]\right)^{q/p}}
    \left.\le \chi^{(p-q)/p}
  \rule{0cm}{0.8cm}\right) \ge 1-2\delta.
  \end{split}
\end{align}From Lemma~\ref{lemma_medianofmean_probability} and by the definition of~$C_{p,q}$, we have
\begin{align}
  \begin{split}\label{C_deflate}
    \mathbb{E}  [\left| Y_i - v_q \right|^{p/q} ] 
    &\le \mathbb{E}  [ Y_i^{p/q} + v_q^{p/q}]
    =\mathbb{E}  \left[ |X_i - \theta_i|^p + v_q ^{p/q}\right]\\
    &=  v_p + v_q ^{p/q} 
    \le   (C_{p,q}+1) v_q^{p/q}.
  \end{split}
\end{align}
Merging the two inequalities~\eqref{probability_two_variances_order} and~\eqref{C_deflate}, with probability at least 
$1-2\delta$, 
we have
\begin{align*}
   \theta_i \le \hat\mu(\{X_{i,j}\}_{j=1}^s, k)+ \left(
    \frac{12 \hat\mu(\{|X_{i,j}-\theta_i|^q\}_{j=1}^s, k)}{
      \max\left\{ 0, 1- C' \chi^{(p-q)/p} \right\}
    }
  \right)^{1/q} \chi^{(q-1)/q}.
\end{align*}
The proof is thus completed by the definition of $\tilde \mu(\{X_{i,j}\}_{j=1}^s, \delta)$.
$\hfill\qedsymbol$

\section{Proof of Corollary ~\ref{corollary_medianofmean}}\label{pf_cor_1}
  This corollary is a direct consequence of Lemma~\ref{lemma_medianofmean_probability} and the inequality in~\eqref{C_deflate}.
  $\hfill\qedsymbol$

\section{Proof of Lemma~\ref{lemma_chosen_time_estimate_extended_robust}}\label{pf_thm_1}
In addition to notations \eqref{short_notes_k}, \eqref{short_notes_chi} and \eqref{short_notes_C}, we introduce two new notations:
\begin{align}
  \tau(x) &:= \frac{1}{1- C' x^{(p-q)/p}},\label{short_notes_tau}\\
  \zeta &:= 12 \tau(\chi) \hat \mu\left( \{|X_{i,j}-\tilde \mu(\{X_{i,j}\}_{j=1}^s, \delta)|^q\},k \right)\label{short_notes_zeta}.
\end{align}
By the definition of 
$\tilde \mu(\{X_{i,j}\}_{j=1}^s, \delta)$, 
we have
\begin{align}\label{relation_tilde_and_hat_mu}
    \tilde \mu(\{X_{i,j}\}_{j=1}^s, \delta) = \hat \mu(\{X_{i,j}\}_{j=1}^s, k)+ \zeta^{1/q} \chi^{(q-1)/q}.
\end{align}
We need to consider the probability of the following event:
\begin{align}\label{need_to_estimate_delta_ineq}
  \tilde \mu(\{X_{i,j}\}_{j=1}^{s}, \delta) - \theta_i \le \Delta_i,
\end{align}
where $\Delta_i$ is defined in \eqref{def_deltai}.
First we estimate an upper bound of 
$\zeta$.
Suppose that
$\tau(\chi) >0$, 
we have
\begin{align*}
  \left| X_{i,j}-\tilde \mu(\{X_{i,j}\}_{j=1}^{s}, \delta) \right|^q \le 2^{q-1} |X_{i,j} - \theta_i|^q + 2^{q-1} |\theta_i - \tilde \mu(\{X_{i,j}\}_{j=1}^{s}, \delta)|^q.
\end{align*}
Recall that 
$Y_{i,j} = |X_{i,j} - \theta_i|^q$ defined in~\eqref{yij_def}.
Thus
\begin{align*}
  \zeta \le& 12 \tau(\chi) \left( 
    2^{q-1} \hat \mu\left(
      \{Y_{i,j}\}_{j=1}^s, k
    \right)
    + 
    2^{q-1} 
    |\theta_i - \tilde \mu(\{X_{i,j}\}_{j=1}^{s}, \delta)|^q
   \right)\\
   \le& 12 \tau(\chi) \left(
    2^{q-1} \hat \mu\left(
      \{Y_{i,j}\}_{j=1}^s, k
    \right)\right.
    \\&\left. + 2^{2q-2}\left(
      |\theta_i - \hat \mu(\{X_{i,j}\}_{j=1}^s, k)|^q
      + |\hat \mu(\{X_{i,j}\}_{j=1}^s, k) - \tilde \mu(\{X_{i,j}\}_{j=1}^{s}, \delta)|^q
    \right)
   \right) \\
   = & 12 \tau(\chi) \left(
    2^{q-1} \hat \mu\left(
      \{Y_{i,j}\}_{j=1}^s, k
    \right)
    + 2^{2q-2}\left(
      |\theta_i - \hat \mu(\{X_{i,j}\}_{j=1}^s, k)|^q
      + \zeta \chi^{q-1}
    \right)
   \right).
\end{align*}
The inequality above uses~\eqref{relation_tilde_and_hat_mu} and also the fact that 
\begin{align*}
  |a+b|^q \le 2^{q-1}(|a|^q + |b|^q)
\end{align*}
by the convexity of $|x|^q$ ($q>1$) and Jensen's inequality~\citep{28}.
Suppose that
\begin{align}\label{chi_bound_1st}
  1>3\cdot 2^{2q} \chi^{q-1} \tau(\chi) ,
\end{align}
then 
\begin{align*}
  \zeta \le \frac{12\tau(\chi) \left( 2^{q-1}\hat \mu(\{ Y_{i,j} \}_{j=1}^s,k) + 2^{2q-2} |\theta_i - \hat \mu(\{X_{i,j}\}_{j=1}^s, k)|^q \right)}{1-3\cdot 2^{2q}\chi^{q-1}\tau(\chi) }.
\end{align*}
Suppose that~$\delta$ and~$s$ meet the conditions in Lemma~\ref{lemma_medianofmean_probability}. Then with probability at least $1-2\delta$, we have
\begin{align}\label{two_direction_abs_bound}
  |\hat \mu(\{X_{i,j}\}_{j=1}^s, k) - \theta_i| \le (12 v_q)^{1/q} \chi^{(q-1)/q}.
\end{align}
Using \eqref{two_direction_abs_bound} and Corollary \ref{corollary_medianofmean}, by defining a function $B(x)$ for simplicity as 
\begin{align}\label{def_bchi}
  B(x) := \frac{6\cdot 2^q \tau(x)  }{1-3 \cdot 2^{2q} x ^ {q-1}\tau(x)}\left(
    1 + C'  x^{(p-q)/p}
  \right)+ \frac{3\cdot 2^{2q} \tau(x) \cdot 12 x^{(q-1)} }{1-3 \cdot 2^{2q} x ^ {q-1}\tau(x)},
\end{align}
then with probability at least $1-3\delta$, we have
\begin{align}\label{BX_estimation}
  \zeta \le v_q B(\chi).
\end{align}

Now we go back to the upper bound estimation of 
$\tilde \mu(\{X_{i,j}\}_{j=1}^{s}, \delta) - \theta_i$.
Using Lemma~\ref{lemma_medianofmean_probability} again and~\eqref{BX_estimation}, with probability at least 
$1-3\delta$, 
we have
\begin{align*}%
    \tilde \mu(\{X_{i,j}\}_{j=1}^{s}, \delta) - \theta_i =& (\tilde \mu(\{X_{i,j}\}_{j=1}^{s}, \delta) - \hat \mu(\{X_{i,j}\}_{j=1}^s, k)) + (\hat \mu(\{X_{i,j}\}_{j=1}^s, k) - \theta_i)\\
    =& \zeta^{1/q}\chi^{(q-1)/q} + (\hat \mu(\{X_{i,j}\}_{j=1}^s, k) - \theta_i)\\
    \le & v_q^{1/q} B(\chi)^{1/q} \chi^{(q-1)/q} + (12 v_q)^{1/q}\chi^{(q-1)/q}\\
    = & v_q^{1/q}  \chi^{(q-1)/q} \left(B(\chi)^{1/q}+12^{1/q} \right).
\end{align*}
Therefore, if 
\begin{align}\label{deflate_1_3delta}
  v_q^{1/q}  \chi^{(q-1)/q} \left(B(\chi)^{1/q}+12^{1/q} \right) \le \Delta_i
\end{align}
holds, then \eqref{need_to_estimate_delta_ineq} is true with probability at least $1-3\delta$.

Again, we need to emphasize the assumptions under which \eqref{need_to_estimate_delta_ineq} is true with probability at least $1-3\delta$:  $\tau(\chi)>0$, \eqref{chi_bound_1st} and \eqref{deflate_1_3delta}.
Equivalently, the following three inequalities are required for $x=\chi$:
\begin{align}
  1 >& C' x^{(p-q)/p},\label{ineq_1}\\
  1 >& 3\cdot 2^{2q} x ^{q-1} \tau(x) ,\label{ineq_2}\\
  \Delta_i\ge&  v_q^{1/q} x^{(q-1)/q} (B(x)^{1/q}+12^{1/q}).\label{ineq_3}
\end{align}
Observe that both $\tau(x)$ and$B(x)$ decrease as $x$ decreases and 
\begin{align*}
  \lim_{x\searrow 0}x^{(q-1)/q} (B(x)^{1/q}+12^{1/q}) = 0.
\end{align*}
So one can always choose a $\chi_M>0$ such that~\eqref{ineq_1}, \eqref{ineq_2} and \eqref{ineq_3} hold for $x \le \chi_M$.

Note that we also need the conditions $\log \delta^{-1}> 1/\epsilon$ and
\begin{align*}
  s \ge (8+\epsilon) \log \delta^{-1} \frac{8\log \delta^{-1}+1}{\epsilon \log \delta^{-1} - 1}
\end{align*}
to apply Lemma~\ref{lemma_medianofmean_probability}.

Next, we bound $\mathbb{E} [s_i(T)]$ for any~$i$ such that $\theta_i<\theta_1$ and prove the logarithmic regret growth by~\eqref{eqn:psdoReg}.
Define
\begin{align*}
  \mathbb{I} \left( X \right) := \begin{cases}
    1, & X \text{ occurs}\\
    0, & X \text{ doesn't occur}
  \end{cases}
\end{align*}
to denote the characteristic function of an event $X$. Let $\delta = t^{-2}$. We use the notation 
$A_t = i$ to indicate that arm~$i$ is chosen at time~$t$. 
For any~$i$ such that $\theta_i \neq \theta_1$, note that $A_t=i$ implies 
$\tilde\mu(\{X_{1,j}\}_{j=1}^{s_1(t-1)}, t^{-2}) \le \tilde\mu(\{X_{i,j}\}_{j=1}^{s_i(t-1)}, t^{-2})$.
We have
\begin{align}
  \begin{split}\label{s_i_divide}
    &s_i(T) = \sum_{t=1}^T \mathbb{I}\left(A_t = i
  \right)\\
  \le& \sum_{t=1}^T \mathbb{I}\left(
    \tilde \mu(\{X_{1,j}\}_{j=1}^{s_1(t-1)}, t^{-2}) \le \theta_1
  \right)
  +\sum_{t=1}^T \mathbb{I}\left(
    \tilde \mu(\{X_{i,j}\}_{j=1}^{s_i(t-1)}, t^{-2}) \ge \theta_1
  \right).
  \end{split}
\end{align}
We will bound the first part of the right-hand side of~\eqref{s_i_divide} by Lemma~\ref{lemma_mutilde_optimal}. Note that $\log \delta^{-1} > 1/\epsilon$ is equivalent to 
$t > \exp\left(\frac1{2\epsilon}\right)$. Write 
\begin{align}\label{def_l0}
  l_0 := \max\left\{\left\lfloor \exp\left(\frac1{2\epsilon}\right) \right\rfloor, \left\lceil (8+\epsilon)\log T^2 \frac{8\log T^2 + 1}{\epsilon \log T^2 -1} \right\rceil\right\}.
\end{align}
Also note that for sufficiently large $T$, the function $\log x^2 \frac{8 \log x^2+1}{\epsilon \log x^2 -1}$ is increasing for $x\in [l_0+1, +\infty)$. We then have
\begin{align*}
  &\sum_{t=1}^T \mathbb{I}\left(
    \tilde \mu(\{X_{1,j}\}_{j=1}^{s_1(t-1)}, t^{-2}) \le \theta_1
  \right)\\
  \le & l_0 +
  \sum_{t=l_0+1}^T \mathbb{I}\left(
    \tilde \mu(\{X_{1,j}\}_{j=1}^{s_1(t-1)}, t^{-2}) \le \theta_1\text{ and } s_1(t-1)\ge (8+\epsilon)\log T^2 \frac{8\log T^2+1}{\epsilon \log T^2 - 1}
  \right)\\
  \le & l_0 +
  \sum_{t=l_0+1}^T \mathbb{I}\left(
    \tilde \mu(\{X_{1,j}\}_{j=1}^{s_1(t-1)}, t^{-2}) \le \theta_1\text{ and } s_1(t-1)\ge (8+\epsilon)\log t^2 \frac{8\log t^2+1}{\epsilon \log t^2 - 1}
  \right).
\end{align*}
By Lemma~\ref{lemma_mutilde_optimal}, we have
\begin{align*}
  \mathbb{E}\left[\sum_{t=1}^T \mathbb{I}\left(
    \tilde \mu(\{X_{1,j}\}_{j=1}^{s_1(t-1)}, t^{-2}) \le \theta_1
  \right)\right] \le l_0 + \sum_{t=1}^\infty \frac{2}{t^2} \le l_0 + \frac{\pi^2}3.
\end{align*}
Now we bound the second part of the right-hand side of \eqref{s_i_divide}. Define
\begin{align}\label{def_l1}
  l_1:= \max\left\{ 
    l_0, \left\lceil \frac{8+\epsilon}{\chi_M} \log T^2\right\rceil
   \right\}.
\end{align}
For sufficiently large $T$, we have
\begin{align}
  \begin{split}\label{estimation_greater_than_optimal_p}
    &\sum_{t=1}^T \mathbb I\left(
    \tilde \mu(\{X_{i,j}\}_{j=1}^{s_i(t-1)}, t^{-2}) \ge \theta_1
  \right) \\
  \le& l_1 + \sum_{t=l_1+1}^T \mathbb I\left(
    \tilde \mu(\{X_{i,j}\}_{j=1}^{s_i(t-1)}, t^{-2}) - \theta_i \ge \Delta_i \text{ and }s_i(t-1) \ge l_1
  \right).
  \end{split}
\end{align}
By the definition of $\chi$ in \eqref{short_notes_chi} and the choice of $\chi_M$, using \eqref{need_to_estimate_delta_ineq}, we have
\begin{align*}
  \mathbb{E}\left[  
    \sum_{t=1}^T \mathbb I\left(
      \tilde \mu(\{X_{i,j}\}_{j=1}^{s_i(t-1)}, t^{-2}) \ge \theta_1
    \right) 
  \right]\le l_1 + \frac{\pi^2}{2}.
\end{align*}
Finally, we have
\begin{align*}
  \mathbb{E}[s_i(T)] \le l_0 + l_1 + \frac{5\pi^2}6.
\end{align*}

For sufficiently large $T$, we can assume that $l_0 = \left\lceil (8+\epsilon)\log T^2 \frac{8\log T^2 + 1}{\epsilon \log T^2 -1} \right\rceil $. Therefore, the above inequality becomes
\begin{align}
  \begin{split}\label{our_regret_of_HT}
    \mathbb{E}[s_i(T)] &\le 
  \left\lceil (8+\epsilon)\log (T^2) \frac{8\log (T^2) + 1}{\epsilon \log (T^2) -1} \right\rceil 
  \\&+ \max\left\{ 
    \left\lceil (8+\epsilon)\log (T^2) \frac{8\log (T^2) + 1}{\epsilon \log (T^2) -1} \right\rceil, \left\lceil \frac{8+\epsilon}{\chi_M} \log (T^2)\right\rceil
   \right\}
   + \frac{5\pi^2}6.
  \end{split}
\end{align}
The proof is then completed by an obvious bound on the right-hand side for sufficiently large~$T$.
$\hfill\qedsymbol$

\section{Proof of Near-Optimal Regret without Knowing~$C_{p,q}$}\label{pf_thm_2}
Recall all three required assumptions which are related to $C_{p,q}$:
\begin{align}
    1 >& C' x^{1-\frac{q}{p}},\label{inequ_1}\\
  1 >& 3\cdot 2^{2q} x ^{q-1} \tau(x) ,\label{inequ_2}\\
  \Delta_i\ge&  v_q^{1/q} x^{(q-1)/q} (B(x)^{1/q}+12^{1/q}).\label{inequ_3}
\end{align}
From the previous deduction, we conclude that $\mathbb{E}[s_{i}(T)]\le l_0+l_1+\frac{5\pi^{2}}{6}$, where $l_0= \max\{\lfloor e^{\frac{1}{2\epsilon}} \rfloor,\lceil (8+\epsilon)\log T^{2} \frac{8\log T^{2}+1}{\epsilon \log T^{2}-1}\rceil\}=O(\log T)$ and $l_1=\max\{l_0,\lceil \frac{8+\epsilon}{\chi_M}\log T^2 \rceil \}$. If we consider the case of replacing $C_{p,q}$ by $f(T)$ as long as~$T$ is large enough, $\chi_M$(an upper bound of $x$) here actually becomes a $T$-relevant value satisfying three inequalities, while the variation does not depend on the index of a specific arm since those requirements are established for all arms to satisfy. Now we only need to figure out the order of $\frac{1}{\chi_M}$ to prove our statement.

For simplicity, we will discuss the three inequalities in turn. Recall that we use $f(T)$ instead of $C_{p,q}$ and focus on the case when $f(T) \ge C_{p,q}$, i.e. $T>t_{0}$ for some~$t_0>0$.

Note that $ \chi_M \textless \frac{1}{[12(f(T)+1)]^\frac{q}{p-q}} \Rightarrow\eqref{inequ_1} $. Therefore, 
\begin{align}
\frac{1}{\chi_M} = O(f(T)^\frac{q}{p-q}). \label{cond_1}
\end{align}
Let $c_0=\Delta_i$, $c_1=v_q^\frac{1}{q}$, $c_2=12^\frac{1}{q}$, $c_3=(9\cdot2^{q+1})^\frac{1}{q}$, $c_4=3\cdot2^q$. We have $\tau(\chi_M)\cdot\chi_M^{q-1}\textless\frac{1}{c_4} \iff \chi_M  \textless[\frac{1}{c_4\cdot\tau(\chi_M)}]^\frac{1}{q-1} \Rightarrow \eqref{inequ_2} $. Since $\tau(\chi_M)>1$ and $c_4>1$, \eqref{inequ_2} holds if 
\begin{align}
\chi_M \textless[\frac{1}{c_4\cdot\tau(\chi_M)}]^\frac{q}{q-1}.\label{cond_2}
\end{align}
After combining conditions given in \eqref{inequ_1} and \eqref{inequ_2}, we see that \eqref{inequ_3} holds if 
\begin{align*}
c_0 \geq c_1\cdot\chi_M^{1-\frac{1}{q}}\left[c_2+c_3(\frac{\tau(\chi_M)}{1-c_4\chi_M^{q-1}\tau(\chi_M)})^\frac{1}{q}\right].
\end{align*}
This indicates that we only require $c_0[1-c_4\chi_M^{q-1}\tau(\chi_M)]^\frac{1}{q}\geq c_1c_2\chi_M^{1-\frac{1}{q}}+c_1c_3\chi_M^{1-\frac{1}{q}}\tau(\chi_M)^\frac{1}{q}$ to achieve \eqref{inequ_3}. A simple analysis shows the left side $\geq c_0[1-c_4\chi_M^{q-1}\tau(\chi_M)]$ while the right side $\leq c_1c_2\chi_M^{1-\frac{1}{q}}+c_1c_3\chi_M^{1-\frac{1}{q}}\tau(\chi_M)$. Therefore, we only require that 
\begin{align*}
c_0 \geq c_0c_4\chi_M^{q-1}\tau(\chi_M)+c_1c_2\chi_M^{1-\frac{1}{q}}+c_1c_3\chi_M^{1-\frac{1}{q}}\tau(\chi_M)\\
= \chi_M^{1-\frac{1}{q}}[c_0c_4\chi_M^{q+\frac{1}{q}-2}\tau(\chi_M)+c_1c_2+c_1c_3\tau(\chi_M)].
\end{align*}
By the fact that $\chi_M \leq 1$, the above inequality which is a stricter version of \eqref{inequ_3} holds if 
\begin{align*}
c_0 \geq \chi_M^{1-\frac{1}{q}}[(c_0c_4+c_1c_3)\tau(\chi_M)+c_1c_2], 
\end{align*}
i.e.
\begin{align}
\chi_M\leq [\frac{c_0}{k\cdot\tau(\chi_M)+b}]^\frac{q}{q-1},
\label{cond_3}
\end{align}
where $k=c_0c_4+c_1c_3, b=c_1c_2$.

Furthermore, by comparing~\eqref{cond_2} with~\eqref{cond_3}, they will simultaneously hold if 
\begin{align*}
\chi_M \leq \left[\frac{\min\{1,c_0\}}{\max\{k+b,c_4\}\cdot\tau(\chi_M)}\right]^\frac{q}{q-1}
=\left[d(1-(12(f(T)+1))^\frac{q}{p}\chi_M^{1-\frac{q}{p})}\right]^\frac{q}{q-1},
\end{align*}
where $d=\frac{\min\{1,c_0\}}{\max\{k+b,c_4\}}$. Since $\chi_M<1$, we only require that
\begin{align}
\chi_M \leq (\frac{d}{1+[12(f(T)+1)]^\frac{q}{p}d})^\frac{1}{D}, ~~~D=1-\max\{\frac{1}{q}, \frac{q}{p}\}>0. \label{cond_4}
\end{align}
Finally, through a comparison between \eqref{cond_1} and \eqref{cond_4}, we have $\frac{1}{\chi_M}=O(f(T)^{\max\{\frac{q}{pD},\frac{q}{p-q}\}})=O(f(T)^\frac{q}{pD})$. By setting $ l_0=\max\{t_0,\lfloor e^{\frac{1}{2\epsilon}} \rfloor,\lceil (8+\epsilon)\log T^{2} \frac{8\log T^{2}+1}{\epsilon \log T^{2}-1}\rceil\}$, we have $l_1=O(f(T)^\frac{q}{pD}\log T)$ and $\mathcal{R}_\pi^\mathcal{F}(T) = O(l_0+l_1+\frac{5\pi^{2}}{6})=O(f(T)^\frac{q}{pD}\log T)$. 
$\hfill\qedsymbol$

\section{Proof of Theorem~\ref{theorem_compare_scalefree}}\label{pf_thm_3}
First note that, since $\Delta_i$ is sufficiently small and $\epsilon$, $C'$ are not related to $\Delta_i$, we only need to show that (noticing that $\epsilon>8$ implies $(8+\epsilon)\frac8\epsilon \le 16$)
  \begin{align}
    (8+\epsilon)\frac1{\chi_M}
    < 3648 \frac{v_2}{\Delta_i^2}.
  \end{align}
  Equivalently, it is sufficient to show that 
  \begin{align}\label{chiM_greater_than_scalefree_comp}
    x':= \frac{(8+\epsilon)\Delta_i^2}{3648 v_2} <\chi_M
  \end{align}
  for sufficiently small $\Delta_i$.
  In order to prove the above, we need to show that \eqref{modified_cond_1}, \eqref{modified_cond_2} and \eqref{modified_cond_3} are all true for $x = x'$ based on the definition of $\chi_M$. 
  For \eqref{modified_cond_1}, let $x = x'$, we have $1> \frac{C' \Delta_i}{8}\sqrt{\frac{\epsilon+8}{57v_2}}$, which is true for small $\Delta_i$. Furthermore, \eqref{modified_cond_2} becomes 
  \begin{align*}
    1 > 6(8+\epsilon)\Delta_i^2 \frac{1}{456v_2 - \sqrt{57}C' \Delta_i \sqrt{(8+\epsilon)v_2}},
  \end{align*}
  which also holds for sufficiently small $\Delta_i$.
  For \eqref{modified_cond_3}, let 
  $x = x'$ 
  and the inequality becomes
  \begin{align}\label{ineq_lastcompare}
    (\epsilon+8) \left(\sqrt{6} \sqrt{
      \frac{v_2 \left(\sqrt{57} (C'+24) \Delta_i \sqrt{\frac{\epsilon+8}{v_2}}+456\right)}{456 v_2-\sqrt{57} C' \Delta_i \sqrt{(\epsilon+8) v_2}-6 \Delta_i^2 (\epsilon+8)}
      }+\sqrt{3}\right)^2 < 912.
  \end{align}
  Taking the limit $\Delta_i \to 0$, \eqref{ineq_lastcompare} becomes 
  $\left(6 \sqrt{2}+9\right) (\epsilon+8)< 912$, 
  which 
  holds for 
  $\epsilon<\frac{280-16 \sqrt{2}}{2 \sqrt{2}+3}$. 
  By continuity, we conclude that \eqref{chiM_greater_than_scalefree_comp} holds for sufficiently small $\Delta_i$ if $\epsilon<\frac{280-16 \sqrt{2}}{2 \sqrt{2}+3}$.
  $\hfill\qedsymbol$

\section{Proof of Theorem~\ref{theorem_lower_bound}}\label{pf_thm_5}
If $\Delta_i$ of arm $i$ is large, that is, if~\eqref{modified_cond_1} and~\eqref{modified_cond_2} directly lead to~\eqref{modified_cond_3}, then $\mathbb{E} [s_i(T)] / \log T$ is upper bounded by a constant not related to $\Delta_i$.
  We are interested in the case where $\Delta_i$ is small, i.e., when~\eqref{modified_cond_3} becomes the main constraint of $\chi_M$:
  \begin{align}
    \Delta_i = \sqrt{v_2 \chi_M} \left(
      \sqrt{B(\chi_M)} + 2\sqrt 3
    \right).
  \end{align}
  By~\eqref{modified_cond_1}, since $B(x)$ is an increasing function of $x$, 
  $B(\chi_M)$ is bounded by a constant only related to $C'$. 
  We have
  \begin{align}
    \frac 1{\chi_M} \le C_{B} \frac{v_2}{\Delta_i^2},
  \end{align}
  where $C_B = B(1/{C'}^2)$ is a constant. 
  Note that
  \begin{align}
    \begin{split}
      \lim\sup \frac{\mathbb{E} [s_i(T)]}{\log T} 
    & \le 
    2(8+\epsilon)
    \left(
      1 + \max\left\{ 
        \frac{8}{\epsilon} , \frac{1}{\chi_M}  
      \right\}
    \right) \\
    & \le 
    2(8+\epsilon)
    \left(
      1 + \max\left\{ 
        \frac{8}{\epsilon} , C_{B} \frac{v_2}{\Delta_i^2} 
      \right\}
    \right).
    \end{split}
  \end{align}
  With the lower bound in the non-parametric setting in \citet{6} and Theorem~\ref{theorem_theoretical_bound}, the proof is finished immediately.
  $\hfill\qedsymbol$

\section{Proof of Theorem~\ref{theorem_ucb1lt}}\label{pf_thm_6}
Define 
\[
  c(t,s) := \begin{cases}
    \sqrt{\frac{a_1 \log t }{s}}, & \sqrt{\frac{a_1 \log t }{s}} < \zeta u_0\\
    \frac{a_2 \log t }{s}, & \sqrt{\frac{a_1 \log t }{s}} \ge \zeta u_0
  \end{cases}.
\]

Similar to the procedure in \citet{3}, for any integer 
$L> 0$ 
and 
$n$ 
such that 
$\theta_i < \theta_1$, 
we have
\begin{align*}
  \mathbb{E} [s_i(T)] \le &L + \sum_{t=1}^T \mathbb P \{
    \overline X_{i,t} + c(t, s_i(t-1)) \ge \overline X_{1,t} + c(t, s_1(t-1))\ \text{and}\ s_i(t-1) > L
  \} \\
  \le& L + \sum_{t=1}^\infty \sum_{s=1}^{t-1} \sum_{k=L}^{t-1}  \mathbb P \left( 
    \frac1k \sum_{j=1}^k X_{i,j} + c(t,k) \ge \frac1s \sum_{j=1}^s X_{1,j} + c(t,s)
  \right)\\
  \le& L + \sum_{t=1}^\infty \sum_{s=1}^{t-1} \sum_{k=L}^{t-1}  
    \mathbb P \left( 
      \frac1k \sum_{j=1}^k X_{i,j}  \ge \theta_i + c(t,k)
      \right)\\
      &+ \mathbb P \left( 
      \frac1s \sum_{j=1}^s X_{1,j}  \le \theta_1 - c(t,s)
      \right)
      + \mathbb P \left( 
      \theta_i + 2 c(t, k) > \theta_1
      \right).
\end{align*}

Choose 
$L_0 = \lceil \max\{ \frac{4a_1 \log T}{(\theta_1 - \theta_i)^2}, \frac{2a_2 \log T}{\theta_1 - \theta_i} \} \rceil$.
We have, $\forall k \ge L_0$,
\begin{align*}
  c(t,k) 
  \le & \max\left\{ \sqrt{\frac{a_1 \log t}{k}}, \frac{a_2 \log t}{k} \right\}
  \le \max\left\{ \sqrt{\frac{a_1 \log t}{L_0}}, \frac{a_2 \log t}{L_0} \right\}\\
  \le & \max\left\{ \sqrt{a_1 \log t \frac{(\theta_1 - \theta_i)^2}{4a_1 \log T}}, a_2 \log t\frac{\theta_1 -\theta_i}{2 a_2 \log T} \right\}\\
  \le & \frac{\theta_1 - \theta_i}{2}.
\end{align*}
Then 
\begin{align*}
  \mathbb{E} [s_i(T)] \le L_0 + \sum_{t=1}^\infty \sum_{s=1}^{t-1} \sum_{k=L_0}^{t-1} 
  \mathbb P \left( 
      \frac1k \sum_{j=1}^k X_{i,j}  \ge \theta_i + c(t,k)
    \right)\\
  + \mathbb P \left( 
      \frac1s \sum_{j=1}^s X_{1,j}  \le \theta_1 - c(t,s)
      \right).
\end{align*}
Now we bound the probabilities by the Bernstein-type bound~\eqref{eqn:BernsteinBound}. 
If 
\begin{align*}
  \sqrt{\frac{a_1 \log t}{k}} < \zeta u_0,
\end{align*}
then
\begin{align*}
  \mathbb P\left( \frac1k \sum_{j=1}^k X_{i,j} \ge \theta_i + c(t,k) \right) 
  = & \mathbb P\left( \frac1k \sum_{j=1}^k X_{i,j} \ge \theta_i + \sqrt{\frac{a_1 \log t}{k}} \right)\\
  \le \exp\left( -\frac{k}{2\zeta} \left( \sqrt{\frac{a_1 \log t}{k}} \right)^2 \right) 
  \le & t^{-4};
\end{align*}
otherwise
\begin{align*}
  \mathbb P\left( \frac1k \sum_{j=1}^k X_{i,j} \ge \theta_i + c(t,k) \right)
  = & \mathbb P\left( \frac1k \sum_{j=1}^k X_{i,j} \ge \theta_i + \frac{a_2 \log t}{k} \right)\\
  \le  \exp\left( -\frac{ku_0}{2}  \frac{a_2 \log t}{k}\right)
  \le & t^{-4}.
\end{align*}
The same argument also applies to 
$\mathbb P \left( \frac1s \sum_{j=1}^s X_{1,j}  \le \theta_1 - c(t,s) \right)$. 
Together we have
\begin{align*}
  \mathbb E [s_i(T)] \le L_0 + 2\sum_{t=1}^\infty \sum_{s=1}^{t-1} \sum_{k=L_0}^{t-1} t^{-4}\le \max\left\{ \frac{4a_1 \log T}{(\theta_1 - \theta_i)^2}, \frac{2a_2 \log T}{\theta_1 - \theta_i} \right\} + 1 +\frac{\pi^2}{3},
\end{align*}
A direct substitution of the above into~\eqref{eqn:psdoReg} completes the proof.
$\hfill\qedsymbol$

\section{Proof of Lemma~\ref{regret_ub_comp_lt}}\label{pf_lem_6}
We only need to show that
  \begin{align}\label{alpsi_regret_lt_regret_comp_to_prove}
     \frac{\alpha x}{\psi^*(x/2)} \le x \max\left\{\frac{4a_1}{x^2}, \frac{2a_2}{x}\right\}
  \end{align}
  for any $x>0$.
  From \eqref{alpha_psi_psi_assumption}, we have
  \begin{align*}
    \frac{\alpha x}{\psi^*(x/2)} = \begin{cases}
      \frac{8\alpha \zeta}{x}, & 0<x<2\zeta u_0\\
      \frac{2\alpha x}{u_0 (x-\zeta u_0)}, & x \ge 2\zeta u_0
    \end{cases}.
  \end{align*}
  From UCB1-LT, we have $a_1 \ge 8 \zeta$ and $a_2 \ge a_1/(\zeta u_0)\ge 8/u_0$, then
  \begin{align*}
    x \max\left\{\frac{4a_1}{x^2}, \frac{2a_2}{x}\right\} 
    \ge \max\left\{\frac{32\zeta}{x}, \frac{16}{u_0} \right\} 
    = \begin{cases}
      \frac{32\zeta}{x}, & 0<x<2\zeta u_0\\
      \frac{16}{u_0}, & x\ge 2\zeta u_0
    \end{cases}.
  \end{align*}
  Since $\frac{8\alpha \zeta}{x} \le \frac{32\zeta}{x}$ as $\alpha \le 4$ 
  and $\frac{2\alpha x}{u_0 (x-\zeta u_0)} \le \frac{16}{u_0}$ as $x\ge 2\zeta u_0$,
  \eqref{alpsi_regret_lt_regret_comp_to_prove} is proved.
  $\hfill\qedsymbol$

\end{document}